\documentclass[10pt,twocolumn,letterpaper]{article}

\usepackage{cvpr}              
\usepackage{multirow}
\usepackage{makecell}
\usepackage{pifont}
\usepackage{xcolor}
\usepackage{colortbl}
\usepackage{pifont}
\usepackage{marvosym}
\usepackage{tabularx,booktabs,array}
\definecolor{cvprblue}{rgb}{0.21,0.49,0.74}
\usepackage[pagebackref,breaklinks,colorlinks,allcolors=cvprblue]{hyperref}

\def\paperID{***} 
\def\confName{CVPR}
\def\confYear{2026}

\title{Scaling Up AI-Generated Image Detection with Generator-Aware Prototypes}

\author{ 
\small Ziheng Qin$^{1,2,3*}$, Yuheng Ji$^{1,3*}$, Renshuai Tao$^{2}$, Yuxuan Tian$^{1,3}$,  Yuyang Liu$^{1,3}$,  Yipu Wang$^{1,4}$, Xiaolong Zheng$^{1,3,4,\text{\Letter}}$ 
\\
$^1$ \small Institute of Automation, Chinese Academy of Sciences \\
$^2$ \small Institute of Information Science, Beijing Jiaotong University \\
$^3$ \small School of Artificial Intelligence, University of Chinese Academy of Sciences \\
$^4$ \small School of Advanced Interdisciplinary Sciences, University of Chinese Academy of Sciences \\
}

\begin{document}
\twocolumn[{
    \maketitle 
    \vspace*{-0.2in} 
    \centering
    \includegraphics[width=1.0\linewidth]{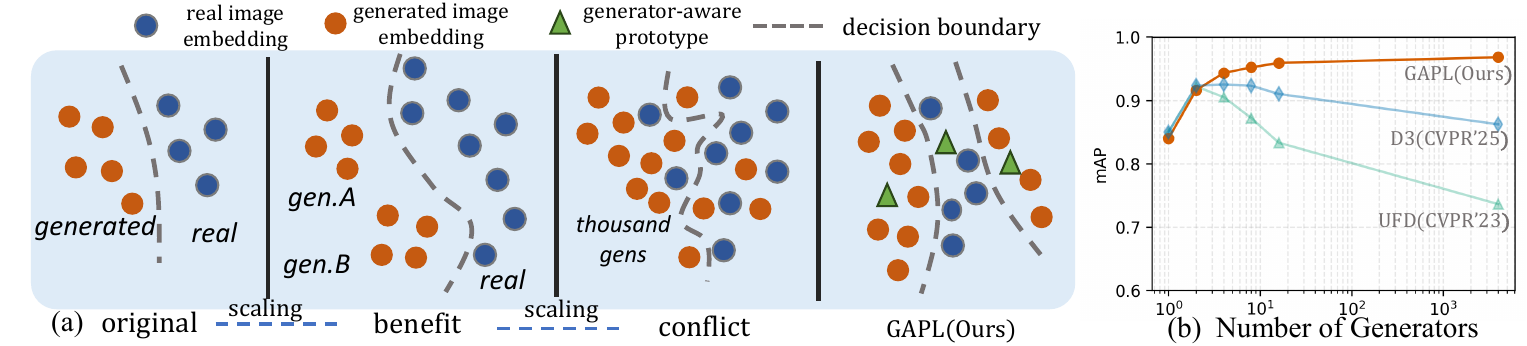}
    \captionof{figure}{Illustration of the ``benefit then conflict" phenomenon we identify. As more generators added to train the detector, it first benefit from the new domain knowledge but eventually degrades due to inseparable embedding. Our proposed method, GAPL, resolves this issue by learning generator-aware prototypes, leading to a substantial mAP improvement over existed detectors in the scaling up settings.}
    \label{fig:teaser} 
    
    \vspace{0.2in} 
}]

\let\thefootnote\relax\footnotetext{$^{*}$ Equal contribution.}
\let\thefootnote\relax\footnotetext{$^{\text{\Letter}}$ Corresponding author.}


\begin{abstract}
The pursuit of a universal AI-generated image (AIGI) detector often relies on aggregating data from numerous generators to improve generalization.
However, this paper identifies a paradoxical phenomenon we term the \textbf{``Benefit then Conflict''} dilemma, where detector performance stagnates and eventually degrades as source diversity expands.
Our systematic analysis, diagnoses this failure by identifying two core issues: severe \textbf{data-level heterogeneity}, which causes the feature distributions of real and synthetic images to increasingly overlap, and a critical \textbf{model-level bottleneck} from fixed, pretrained encoders that cannot adapt to the rising complexity.
To address these challenges, we propose Generator-Aware Prototype Learning (GAPL) , a framework that constrain representation with a structured learning paradigm.
GAPL learns a compact set of canonical forgery prototypes to create a unified, low-variance feature space, effectively countering data heterogeneity.
To resolve the model bottleneck, it employs a two-stage training scheme with Low-Rank Adaptation, enhancing its discriminative power while preserving valuable pretrained knowledge.
This approach establishes a more robust and generalizable decision boundary. Through extensive experiments, we demonstrate that GAPL achieves state-of-the-art performance, showing superior detection accuracy across a wide variety of GAN and diffusion-based generators. Code is available at \url{https://github.com/UltraCapture/GAPL}
\end{abstract}

\section{Introduction}
\label{sec:intro}

The increasing sophistication of generative models \cite{rombach2022high, AdobeFirefly, wu2025qwenimage} necessitates the development of reliable AI-generated image (AIGI) detectors to mitigate risks such as misinformation. An ideal detector must generalize to generated images from a wide array of generators. To this end, a common strategy is to expand the training dataset by aggregating data from multiple sources \cite{yang2025d3, park2025commfor} to turn the common \textit{Train on one} to \textit{Train on many}.

\begin{figure}
    \centering
    \includegraphics[width=1.0\linewidth]{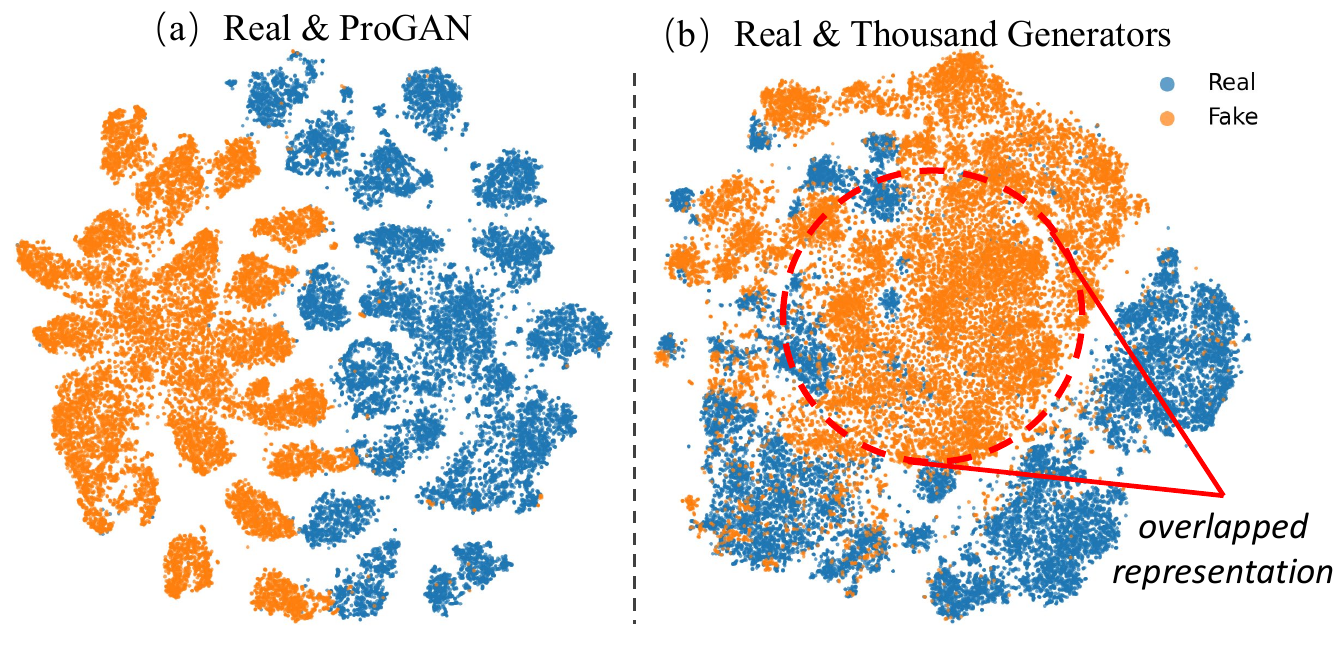}
    \caption{\textbf{T-SNE visualization on CLIP Embeddings.} \textbf{(a)} Comparison between images generated by a single generator and real images. \textbf{(b)} Comparison between images generated by thousands of diverse generators and real images.}
    \label{fig:tsne}
\end{figure}

However, in the process of directly scaling up existing AIGI detector \cite{ojha2023towards,yang2025d3}, performance does not always increase as we wish. Instead, we observe a paradoxical phenomenon of \textit{\textbf{Benefit then Conflict}} in scaling up experiments; as illustrated in Fig.~\ref{fig:teaser} while aggregating data from a few generators is beneficial, the detector's efficacy diminishes as the diversity of training sources expands.  This counterintuitive outcome reveals fundamental limitations in scaling current AIGI detectors. To understand the underlying reason for this phenomenon, we construct a series of datasets equal number of generated images, but varying numbers of generators. Then we use a Linear Discriminant Analysis (LDA) model to assess the intrinsic separability in existing detectors, revealing two conclusions: (1) scaling itself causes data towards inseparable, (2) training in an end to end manner, which thought usually to be overfitting in domain, outperforms pretrained-based ones in scaling up setting. Based on these findings, we identify two primary challenges: \textit{\textbf{(1) Data-Level Heterogeneity}}: The feature distributions of images from different generators are highly diverse thereby amplifying the intrinsic difficulty of the classification task. As shown in Fig.~\ref{fig:teaser}(a), this inseparability cause a irregular decision boundary in detectors.  \textit{\textbf{(2) Model-Level Bottleneck}}: Many state-of-the-art detectors \cite{ojha2023towards, liu2024forgery, tan2025c2p, yan2024sanity, wen2025spot} rely on fixed, pre-trained encoders (e.g., CLIP \cite{CLIP})  for feature extraction. While these models provide powerful semantic priors to achieve generalization when training on single source, these priors may be contradictory when learning from heterogeneous data. As shown in Fig.~\ref{fig:tsne}, with only one generator, the fixed CLIP embedding space shows a clear decision boundary. However, when the number of generators increases to thousands, the features exhibit significant embedding overlap.

To address these challenges, we propose \textbf{Generator-Aware Prototype Learning (GAPL)}, following a structured \textit{Turn Thousands into a Few} philosophy. Rather than treating all forgery sources equally, GAPL learns a compact representation of forgery. To handle \textit{data-level heterogeneity}, it learns a small set of prototypes from a few canonical generators, where each prototype captures a canonical forgery pattern. The \textbf{Prototype Mapping (PM)} mechanism reconstructs image embeddings as a linear combination of prototypes weighted by similarity, constraining feature variation within the prototype-defined space. To address the \textit{model-level bottleneck}, GAPL adopts a two-stage training scheme that injects forgery cues while preserving the pretrained encoder’s generalization. In Stage 1, an MLP is trained to make the feature space forgery-aware; in Stage 2, LoRA \cite{hu2022lora} tunes a low-rank subspace that works synergistically with the PM mechanism. Across six benchmarks, GAPL achieves 90.4\% average accuracy and 94.9\% average precision, outperforming the previous state of the art by 3.5\%. Moreover, in-depth experiments demonstrate the excellent robustness of our detector against post processing, thoroughly highlighting its applicability in real world.

Our contributions are summarized as follows:
\begin{itemize}
\item We identify and analyze the ``Benefit then Conflict'' paradox in scaling up setting, revealing the core challenges that impede the scalability of current AIGI detectors.
\item We introduce Generator-Aware Prototype Learning (GAPL), a novel framework that effectively manages the high heterogeneity of AI-Generated images by learning a shared space of forgery prototypes.
\item Our method overcomes the limitations of fixed feature extractors by explicitly training a model to map diverse artifacts to a set of canonical forgery concepts, enabling a more robust and well-defined decision boundary.
\item Through extensive experiments, we demonstrate that GAPL achieves state-of-the-art performance, showing a consistent detection across benchmark with a mean accuracy of 90.4\% across 55 test subsets in 6 benchmarks, outperforms existing detectors.
\end{itemize}

\section{Related Works}
\subsection{Generalized AI-Generated Image Detection}
    Generalized AIGI detection seeks to train a model that can generalize to unseen generative model from a single training source. The primary approach can be divided into two categories. The first category is to extract a low-level artifact pattern, and detection is via recognizing this artifact pattern. This generalized pattern is from different domains, including data augmentation \cite{wang2020cnn} and local pixel dependency \cite{tan2024rethinking, liang2025ferretnet} in the common RGB domain. In frequency domain, high frequency component is widely used \cite{tan2024frequency, li2025improving, bihpf}, other domains like reconstruction error \cite{wang2023dire, luo2024lare} and VAE introduced artifact \cite{chen2025dda,rajan2025aligned, liu2025beyond} prove that AIGI artifacts widely exist. The second is through semantic-level artifact, based on the fact that generative models often fail to maintain semantic consistency, which can be detected via a pretrained image encoder. UniFD \cite{ojha2023towards} first explored this direction using a frozen CLIP image encoder. Following works exploit the ability of pretrained encoders in AIGI detection via inserting adapter in CLIP \cite{liu2024forgery, li2025towards} , creating tunable subspace \cite{yan2024orthogonal, zhang2025towards, liu2026mirror}, injecting forgery concept \cite{tan2025c2p} and extract hybrid feature for detection \cite{cheng2025co, yan2024sanity}. However, these detectors often fail to maintain consistency performance under domain gap due to the fingerprint or inconsistency they aim to extract does not appear in all generative models.

\subsection{Curated datasets for AIGI Detection}

    Existed AIGI detectors were trained on a paired dataset, usually comprise of only one generator or one with its variant \cite{wang2020cnn,chen2024drct, Guillaro2024biasfree,chen2025dda}. To address the generalization challenges, some efforts have begun attempting the use of data from a more diverse source \cite{yang2025d3, yan2024sanity, wen2025spot} to build a more convincing detector. However, their approach to curate the dataset is just simply concatenate the existing datasets, which still lacks breadth and diversity. On the other hand, a series of work recognizes the importance of generator diversity, focusing on building large-scale datasets. Specifically, RED series \cite{red116, red140} curated datasets from hundreds of generators, while Community Forensics \cite{park2025commfor} expanded this to the thousand-generator level. These works still leaves a gap: they are either aimed at tasks other than AIGI detection \cite{red116, red140}, or they train a detector \cite{park2025commfor} with common vision architecture without specialized design to effectively utilize heterogeneous data. To our knowledge, we are the first to simultaneously consider both model-level and data-level challenges in AIGI detection. 
   
\section{Scaling-up AIGI Detectors}

    In this section, we will delve into the paradox of \textbf{\textit{Benefit then conflict}}, explain how it occurs in scaling up settings in a qualitative way and introduce how the proposed GAPL solves the challenges. 
    
\subsection{Challenges in Scaling Up Setting}  
    \label{}
    A common AIGI detector works in the following way. We denote a paired dataset $\mathcal{D}=\{\mathcal{X}, \mathcal{Y}\}$ , where $\mathcal{X}$ is the images set $\mathcal{X}=\{ x_{1}, x_2, ...x_n\}, x_i \in \mathbb{R}^{3 \times H \times W} $ with corresponding label $\mathcal{Y}=\{y_0, y_1, ..., y_n \}, y_i=\{ 0,1 \}$.
    
    To avoid overfitting, common approach \cite{ojha2023towards, yang2025d3} uses an pretrained encoder to extract image embedding $f_i = \phi(x_i) \in \mathbb{R}^D$ , where $ \phi( \cdot ) $ denotes image encoder, and the image embedding refer to its [CLS] token. Subsequently, a classifier $\mathcal{G}$ is applied to predict its log-likelihood, with its parameter $\theta$ optimized via cross entropy loss on dataset $D$:
    \begin{equation}
    \begin{aligned}
       \hat{y} =  & \mathcal{G}_\theta \left( f \right) ,  \\
       \theta = \mathop{\arg\min}_\theta 
       \mathbb{E}_{(x, y) \in D} \Big[
          - \big( & y \log(\hat{y}) + (1-y)\log(1-\hat{y}) \big)
       \Big].
    \end{aligned}
    \end{equation}

    \textbf{Scaling up data itself causes increasing heterogeneity}. The discrepancy in the capacity of generative models to approximate the true distribution leads to an increased variance in the combined distribution of different generative models. We consider a dataset distributed according to a gaussian mixture, a series of generative models fit this distribution by maximizing likelihood on this dataset: 
    \begin{equation}
        \begin{aligned}
            \{x|(x,0)\} \sim P_{real} &=\sum_{k}^{K}\pi_k \mathcal{N}(\mu_k, \Sigma_k) \\
            \{x|(x,1)\} \sim P_{gen} &=\sum_{i}^{G}w_i\sum_{j}^{K}\pi_{i,j}\mathcal{N}(\mu_{i,j},\Sigma_{i,j}),
        \end{aligned}
    \end{equation}
    then, according to the law of total variance, the covariance matrix can be calculated with:
    \begin{equation}
        \begin{aligned}
            \Sigma_{real} & = \mathbb{E}(\Sigma)+\text{Var}(\mu), \\
            \Sigma_{gen} & = \mathbb{E}_G \left[ \text{Var}(X |G) \right] + \text{Var} _G\left[ \mathbb{E}(X|G)\right] \\
             &  = \underbrace{\mathbb{E}_M \left[ \text{Var}(X|G_i,M)|G_i\right]+\text{Var}_M\left[\mathbb{E}(X|G_i,M)|G_i \right]}_{\text{generator fitting variance}} \\
             & + \underbrace{\text{Var}_G \left[ \mathbb{E}(X|G)\right]}_{\text{cross generator variance}},
        \end{aligned}
    \end{equation}
    where $M$ denotes mode in GMM. The first two terms reflect variance in the learned distribution, which correspond to the terms of the real covariance. The third term, reflects cross generator variance which grows as generator diverse.
    
    To validate our proposal, we conducted an experiment where we collected a series of datasets from GenImage \cite{zhu2024genimage}. In each dataset, there are a total number of 8,000 images from ImageNet \cite{imagenet} and the same number of images generated with several models trained on ImageNet, with the number of models varying from 1 to 8. We also uses a dataset with thousands of generative models to further provide large scale setting. For detailed information of these datasets and our experiment, please refer to Appendix.\ref{sec:pre-exp}.   To estimate the heterogeneity in the dataset, we use scatter matrix of image feature as a proxy for covariance matrix $\Sigma$. For a set $C$, its scatter matrix is calculated by:
    
    \begin{equation}
        S = \sum_{f \in C} \left( f-\mu \right) \left( f-\mu\right)^T,
    \end{equation}
    where $\mu=\frac{1}{N}\sum f$ is the mean feature of the extracted features in this set. We calculate the trace of this scatter matrix $tr(S)$ to evaluate total variance of real and generated images.

\begin{figure}[!tp]
        \centering
        \includegraphics[width=\linewidth]{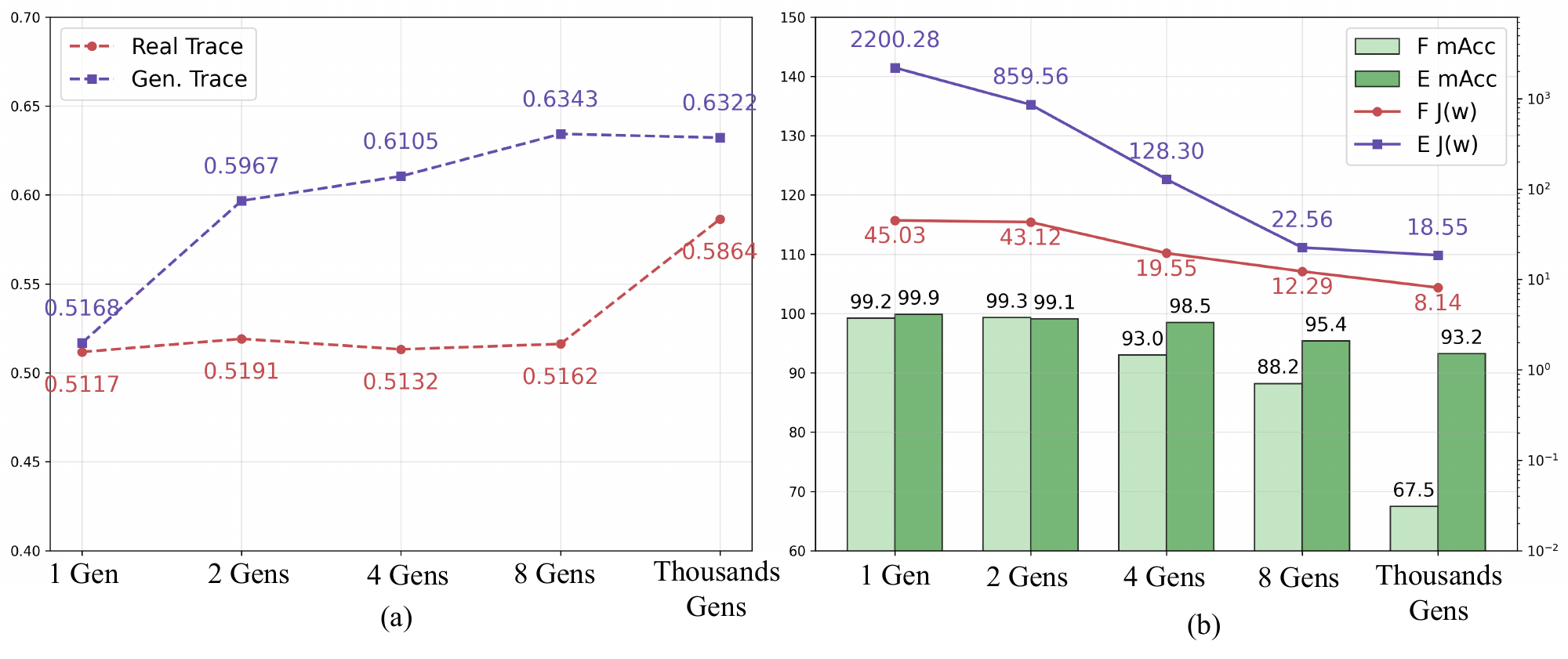}
        \caption{Results of the experiment on a series of datasets comprise of different number of generators.  \textbf{(a)} Trace of the scatter matrix from generated images is significantly higher than that from real images.  \textbf{(b)}  Accuracy of validation and \textit{Fisher ratio $J$} between end-to-end ($\mathbf{E}$) and frozen-encoder models ($\mathbf{F}$), end-to-end models tend to have a higher separability and accuracy, especially in scaling up settings. ``Gen(s)'' denotes ``Generator(s)''. } 
        \label{fig:stat}
    \end{figure}

    Results are presented Fig.~\ref{fig:stat} (a). The figure illustrates that: for the first four datasets, the variance of the generated images exhibits a clear upward trend that scales with the number of generators. Conversely,  the variance of real images is highly consistent. In the last dataset, which incorporates real and generated images from diverse sources,  the variance of the generated distribution is still markedly higher than that of the real distribution. This confirms our hypothesis: the heterogeneity of real images remain stable whereas the heterogeneity of generated images increases as more generators are introduced. 

    \textbf{Reliance on a frozen encoder sets bottleneck to model performance}.
        Common AIGI detectors leverage a pre-trained encoder to avoid overfitting and enhance generalization, This dependency indicates that when training images on novel generators, the model can only push their representations towards the real or generated classes in the feature space, rather than learning the new generative artifacts. This often results in a poor decision boundary and lower discriminability. To investigate the intrinsic discriminability of the feature space, we analyze the embeddings using Linear Discriminant Analysis (LDA) \cite{fisher1936use}. LDA seeks to create a linear projection space and provides  a quantitative measure \textit{Fisher ratio} of this class separability, allowing us to assess how well the pre-trained embeddings can inherently distinguish between real and generated images. Moreover, its optimization can be given in closed form:
        
    \begin{equation}
        \begin{split}
          w & = \mathop{\arg\max}_{w}J (w)=\frac{w^TS_bw}{w^TS_ww} \\
            & =  kS_{w}^{-1}\left( \mu_0-\mu_1 \right),
        \end{split}
    \label{eq:ldaoptim}
    \end{equation}
    where $S_b, S_w$ are scatter between classes and within class matrices, respectively, which can be calculated by
    \begin{equation}
        \begin{split}        
            & S_w = \sum_{i=0}^{C}S_i \\ 
            & S_b = (\mu_0-\mu_1)(\mu_0-\mu_1)^T
        \end{split}
    \label{eq:ldas}
    \end{equation}
        We designed an experiment with the datasets introduced above in the following steps: (1) For each dataset, train an end-to-end classifier, simultaneously, train a classification head based on an frozen image encoder of CLIP. (2) Extract penultimate representation from the two models on validation set, estimate $J$ based on eqs.~\eqref{eq:ldaoptim} and \eqref{eq:ldas} and classify them with a threshold of 0.5. We report detection accuracy and \textit{Fisher ratio} $J$ as metrics.

        The results are presented in Fig.~\ref{fig:stat}(b). Results reveal that first,  \textit{Fisher ratio} and detection accuracy drops as more diverse data adds. Second, an end-to-end detector has a higher \textit{Fisher ratio} on the same data scale compared to the pretrained-based one. Although they achieve good generalization by training on single source with a relatively small \textit{Fisher ratio}, they seemed to be indistinguishable to more generators. This lead us to reflect that the reliance of pretrained encoder in prevalent AIGI detectors may fundamentally impose an upper bound on achievable performance when scaling up AIGI detection.

\subsection{Scaling Up with Generator-Aware Prototypes}
    \begin{figure*}[!tp]
        \centering
        \includegraphics[width=1.0\linewidth]{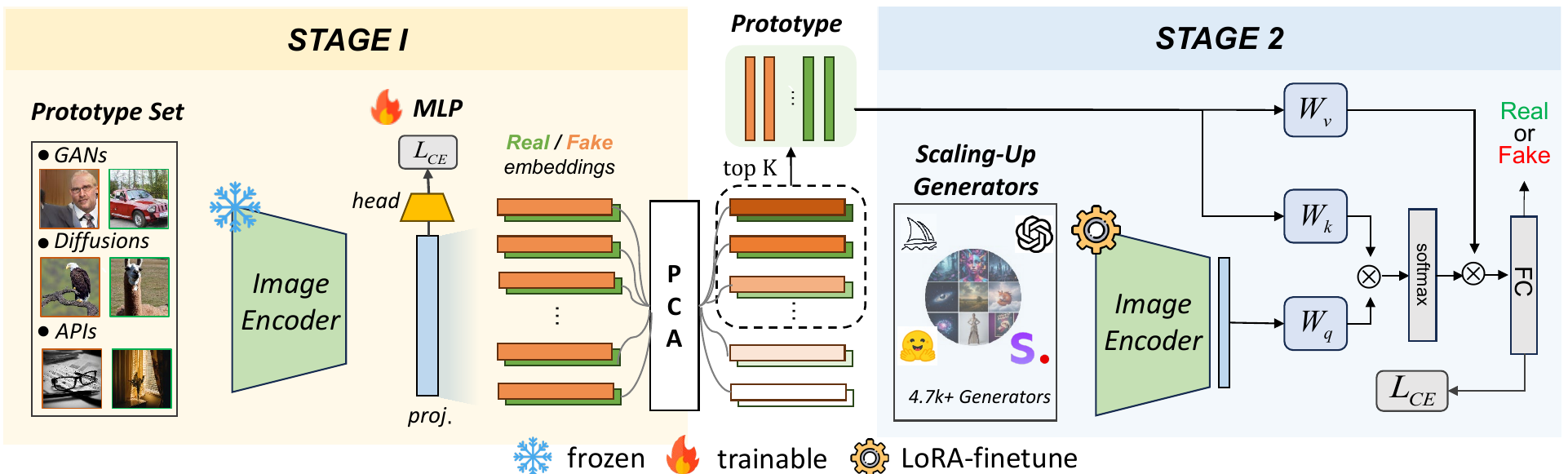}
        \caption{\textbf{The overall framework of our GAPL.} We train the GAPL in two stages. In the first stage, we train a MLP and extract embeddings for PCA decomposition, we only retain top $N$ components as prototypes; in the second stage, we uses LoRA to finetune the encoder and map image embedding to prototypes to get the final logits. }
        \label{fig:method}
    \end{figure*}
    Up to now, we have identified key challenges in building a scaling up AIGI detector:  (1) To lower inner heterogeneity of generated images, we need to achieve a compact representation for the generated images. (2) Fixed embedding space should be abandoned. Relying on a static embedding space like a frozen CLIP is fundamentally limiting. 
    
    To tackle the two challenges, we propose GAPL for scaling up AIGI detection. we adopt a philosophy called \textit{Turn Thousand into a Few}, the GAPL learns a compact shared prototype space to distill thousands of generators into a few, representative prototypes. This process is via a two-stage learning framework, we first leverage pretrained encoder to construct prototype, then we let prototypes guide the encoder to learn and project to this prototype space. The framework of  GAPL is shown in Fig.~\ref{fig:method}.

\textbf{STAGE I: Creating Forensic-Realated Space and Generator-Aware Prototypes.}
   
    The objective of this stage is to perform a supervised projection, creating a forgery-related subspace based on the pretrained embedding space. Since most pretrained encoder are highly correlated with semantic information due to its pretraining objective, we aim to endow the encoder with a basic understanding of generated images, enabling it to capture representative concepts. 
    
    Specifically, We select an amount of $M$ images each from ProGAN\cite{karras2018progressive}, Stable Diffusion v1.4 \cite{rombach2022high}, and Midjourney \cite{Midjourney}, sourced from the ForenSynths \cite{wang2020cnn} and GenImage \cite{zhu2024genimage} as they represent GANs, Latent diffusion models and Commercial APIs to form a prototype set. We also pair the same amount of real images accordingly. Then, an MLP was attached after the encoder to project dimension and get logits for classification, the projection and MLP can be formulated with $g_{proj}: \mathbb{R}^{D} \xrightarrow{} \mathbb{R}^{D'}$ and $\texttt{MLP}: \mathbb{R}^{D} \xrightarrow{} \mathbb{R} $. This process can be formulated as follow:
    \begin{equation}
        \begin{split}
            & f = \phi(x),  \\
            & \hat{y} =\texttt{MLP}\left(\texttt{Normalize}(f) \right),
        \end{split}
    \end{equation}
    where $\phi(\cdot)$ is the image encoder. In the first stage, the image encoder keeps frozen, we only train this MLP with binary crossentropy as loss function on the prototype set.
    
    Subsequently, we utilize the intermediate layer of the MLP to extract forgery-related embeddings to build prototypes. $\mathbf{F}_f = [f_1^f, f_2^f,  ..., f_{M \times 3}^f]$ , $\mathbf{F}_r = [f_1^r, f_2^r,  ..., f_{M \times 3}^r]$, $f \in \mathbb{R}^{D'}$, denotes the real and generated subset of these embeddings. We then perform principle component analysis (PCA) separately on the two subsets for prototype constructing, and select the top $N/2$ principal components from each subset as Generator-Aware Prototypes:
    \begin{equation}
    \begin{split}
            C = \frac{1}{3M - 1} \left( \mathbf{F}_f - \mathbf{1}\mathbf{\bar{\mu}} \right)^T &\left( \mathbf{F}_f - \mathbf{1}\mathbf{\bar{\mu}} \right), \\
            Cv_i=\lambda_iv_i, \\
        \mathbf{P_f}= \left[ v_1, v_2, ..., v_{N/2} \right]^T & \in \mathbb{R}^{(N/2) \times D'},
    \end{split}
    \end{equation}
    where $\bar{\mu} \in \mathbb{R}^{1 \times D'}$ is mean vector of the embeddings in $\mathbf{F}$. $\mathbf{1}^T_{3M} \in \mathbb{R}^{3M}$ denotes all-one vector. We concatenate real and generated prototypes to form the final prototype matrix $\mathbf{P}=[\mathbf{P_r}; \mathbf{P_f}] \in \mathbb{R}^{N \times D'} \ $.
    
    We adopt this selection strategy based on the assumption that, PCA decomposes the covariance matrix, the components with larger variance capture more general forgery-related information, while those with smaller variance reflect generator-specific characteristics. The remaining components, which contribute little to the overall variance, are mainly task-irrelevant noise and thus discarded. 
    
\textbf{STAGE II: Map Generators to Prototypes for Large Scale AIGI Detection.}
   In the second stage, we further enhance the model's capability, adhering to our principle of not relying on a fixed embedding space. To maintain as much knowledge as possible in pretrained image encoder, this adaptation is performed using LoRA \cite{hu2022lora}.
   
    \begin{equation}
        f=g_{proj}\left( \phi_{lora}(x) \right).
    \end{equation}
        
  Furthermore, we aim to solve how to effectively integrate the extracted prototype with the image feature from image encoder. Given that our primary objective is to enforce feature in a low variance way, we posit that a good approach is to represent each image's features into a dynamic linear combination of these learned prototypes. It should create a map from the diverse feature space onto a low-variance prototype space. Therefore, we adopt a simple yet effective cross-attention \cite{vaswani2017attention} mechanism as this mapping to perform feature fusion, where the image embedding act as queries and the prototypes serve as keys and values. It can be expressed as follows: 
\begin{equation}
 \begin{split}
    \tilde{f} & = \texttt{Attn} \left(W_qf, W_k\mathbf{P}, W_v\textbf{P} \right) \\
      & =  softmax(\frac{(fW_q)(\mathbf{P}W_k)^T}{\sqrt{D'}}) \cdot \mathbf{P}W_v 
 \end{split}
 \end{equation}

    Where $W_q,W_k,W_v \in \mathbb{R}^{D' \times D'}$ is the respective affine matrix. We feed this re-organized feature $ \tilde{f} $ into a linear classifier to predict its final logits and train the whole model with binary crossentropy.

\begin{table*}[h]
\centering
\renewcommand{\arraystretch}{1.1}
\resizebox{\textwidth}{!}{
\begin{tabular}{l l cc cc cc cc cc cc cc}
\toprule
\multirow{2}{*}{\textbf{Method}} & \multirow{2}{*}{\makecell{\textbf{Training}\\\textbf{Class/Num.}}} &
\multicolumn{2}{c}{\textbf{ForenSynths}} & \multicolumn{2}{c}{\textbf{UFD}} & \multicolumn{2}{c}{\textbf{GenImage}}  & \multicolumn{2}{c}{\textbf{SynthBuster}} & \multicolumn{2}{c}{\textbf{Chameleon}} & \multicolumn{2}{c}{\textbf{CommFor}} & \multicolumn{2}{c}{\textbf{Mean}} \\
\cmidrule(lr){3-4} \cmidrule(lr){5-6} \cmidrule(lr){7-8} \cmidrule(lr){9-10} \cmidrule(lr){11-12} \cmidrule(lr){13-14} \cmidrule(lr){15-16}
 &  & Acc & AP & Acc & AP & Acc & AP & Acc & AP & Acc & AP & Acc & AP & Acc & AP \\
\midrule
\multicolumn{16}{l}{\cellcolor{gray!25}\textit{GAN-Generalized}} \\
\midrule
CNNDet \cite{wang2020cnn} & ProGAN/720k & 75.3 & 91.8 & 53.1 & 71.1 & 51.3 & 65.6 & 50.3 &  53.9 & 57.0 & 42.6  & 51.5 & 53.7 & 56.4 & 63.1 \\
NPR \cite{tan2024rethinking}  & ProGAN/144k & 90.8 & 91.8 & 95.1 & 97.4 & 86.2 & 91.6 & 38.9 & 45.4 & 57.4 & 47.7 & 73.8 & 74.7 & 73.7 & 74.8 \\
UniFD \cite{ojha2023towards} & ProGAN/720k & 89.4 & 96.7 & 82.2 & 95.7 & 70.0 & 88.3 & 67.6 & 78.0 & 57.2 & 46.2 & 54.0 & 58.3 & 70.1 & 77.2  \\
SAFE \cite{li2025improving} & ProGAN/144k & \underline{96.2} & \underline{98.6} & 95.7 & 99.1 &  95.5 & 99.3 & 44.7 & 44.7 & 59.2 & 50.6 & 50.0 & 48.3 & 73.6 & 79.2 \\
AIDE \cite{yan2024sanity} & ProGAN/720k & 88.4 & 96.0 & 96.7 & 99.0 & 92.0 & 97.5 & 48.0 & 49.9 & 58.4 & 50.1 & 49.9 & 47.1  & 72.2 & 73.3 \\
\midrule
\multicolumn{16}{l}{\cellcolor{gray!25}\textit{Diffusion-Generalized}} \\
\midrule
DRCT \cite{chen2024drct} & SD2.1/473k & 50.0 & 49.7 & 70.0 & 79.2 & 78.0 & 94.0 & 77.6 & 84.7 & \textbf{79.8} & \textbf{85.2} & 49.6 & 49.4 & 67.5 & 73.7 \\ 
Co-SPY \cite{cheng2025co} & SD1.4/320k & 65.6 & 75.9 & 76.8 & 84.3 & 78.0 & 94.1 & 77.4 & 91.6 & 68.8 & 68.8 & 67.9 & 79.2 & 72.4 & 82.3   \\
B-Free \cite{Guillaro2024biasfree} & SD2.1/360k & 81.1 & 94.2 & 87.1 & 95.3 &  87.3 & 97.7 & \textbf{94.9} & \textbf{98.8} & \underline{78.2} & \underline{85.2} & 81.5 & 91.8 & 85.0 & \underline{93.8} \\
\midrule
\multicolumn{16}{l}{\cellcolor{gray!25}\textit{Scaling-Ups}} \\
\midrule
AIDE \cite{yan2024sanity} & 8gens/1.3M & 85.3 & 93.8 & \textbf{98.5} & \textbf{99.9} & \textbf{99.7} & \textbf{100.0} & 52.7 & 56.0 & 65.8 & 69.7 & 50.1 & 48.2 & 75.4 & 77.9 \\
D3 \cite{yang2025d3} & 9gens/2M & 93.0 & 97.9 & 94.8 & 99.0 & 95.4 & 99.2 & 81.3 & 89.5 & 61.0 & 56.5 & 73.6 & 86.0 & 83.2 & 88.0 \\
CommForen \cite{park2025commfor} & 4.7Kgens/5M & 92.3 & 98.2 & 94.0 & 97.0 & 84.0 & 93.4 &  87.0 & 94.7 & 77.5 & 83.5 & \underline{86.8} & \underline{93.9} & \underline{86.9} & 93.4 \\
\rowcolor{cvprblue!20}
GAPL(Ours) & 4.7Kgens/550k & \textbf{97.2} & \textbf{99.5} & \underline{97.2} & \underline{99.8} & \underline{96.7} & \underline{99.6} & \underline{91.1} & \underline{97.2} & 71.0 & 75.6 & \textbf{89.4} & \textbf{97.8} & \textbf{90.4} & \textbf{94.9} \\ 
\bottomrule
\end{tabular}
}
\caption{Overall performance of different detectors across benchmarks. \textit{GAN-Generalized} means that these models were trained on GAN-generated images, \textit{Diffusion-Generalized} means diffsuion-generated images accordingly. \textit{Scaling-ups} means that these model have the access to all kinds of images. We also report each detector's generator type and number of images of their training set.}
\label{exp1}
\end{table*}

\section{Experiments}

In this section, we evaluate the proposed GAPL to validate the followng \textbf{Reseach Questions (RQs)} :
\begin{itemize}
    \item RQ.1: How well does the proposed \textbf{GAPL} in detecting generated images from any generator?
    \item RQ.2: How well is the inherent potential of proposed \textbf{GAPL} in model architecture?
    \item RQ.3: Why is the designed Prototype Learning and two-stage training useful?
    \item RQ.4: How does the number of prototypes and selection of prototype sets affect performance?
    \item RQ.5: How well is the proposed \textbf{GAPL's} robustness in handling post processing?
\end{itemize}

\subsection{Experiment Details}
    \textbf{Training set.} In our GAPL, prototype set and scaling up set are used for training. In the prototype set used in first stage, we select $M = 2000$ images generated by ProGAN \cite{karras2018progressive}, SDv1.4 \cite{rombach2022high} and Midjourney \cite{Midjourney}. These images are sampled from Forensynths and GenImage. As for scaling up dataset, we choose the Community-Forensic \cite{park2025commfor}. This dataset manually chose 12 GANs, 3 pixel diffusion models and collected about 4000 latent diffusion models on huggingface, constituting a large and diverse datasets that covers most generative architectures. We use a small version, compared to the original dataset of 5.4M images, the small version only contains 550k images while the total number of generative models remains the same.

 \begin{table}[t]
        \centering
        \resizebox{0.48\textwidth}{!}{
        \begin{tabular}{l |c c c}
             \toprule
             Dataset & \#Real/\#Fake & \# Models & Type of models  \\
             \midrule
             ForenSynths \cite{wang2020cnn} & 31k/31k & 8 & GAN \\
             UniversalFakeDetect \cite{ojha2023towards} & 8k/8k &8 & Diffusion \\
             GenImage \cite{zhu2024genimage} & 48k/48k & 8 & GAN\&Diffusion \\
             Synthbuster \cite{synthbuster} & 1k/9k & 9 & Diffusion \\
             Chamelon \cite{yan2024sanity} & 14.9k/11.2k & -  & Unknown \\
             Community-Forensics Eval \cite{park2025commfor} & 25k/25k & 21 & Diffusion \\
             \bottomrule
            \end{tabular}
        }
        \caption{The selected 6 AIGI benchmarks to evaluate the methods.}
        \label{tab:testset}
    \end{table}
    
    \textbf{Testing set.} We select six benchmarks to evaluate our method's generalization, covering a spectrum of common and advanced generative models developed over the last decade. Tab.~\ref{tab:testset} lists some information about the benchmarks. Note that despite the use of a large scale training set, more than half of generative models in testing set are still unseen by the detector. See Appendix.~\ref{sec:benchmark} for detailed introduction of these benchmarks.

\textbf{Implementation Details.}
    We use the image encoder of CLIP-ViT:L \cite{CLIP, dosovitskiy2020vit} for finetuning. We select $M=2000$ images each from 3 generators as prototype set, along with the same amount of real images. The dimension of projection is set to be $D'=128$, and the number of prototypes is $N=64$.  The experiment was conducted on 2 GeForce RTX 4090 GPUs. More details are in Appendix.~\ref{sec:impledet}

\subsection{Comprehensive evaluation of detection performance across generators (RQ.1)}
    To evaluate the overall detection performance of the proposed GAPL, we conduct test on the 6 selected benchmarks. The results are shown in Tab.~\ref{exp1}. The results reveal the following: \textit{\textbf{(1) Generalized AIGI detectors fail to maintain performance}}. Generalized detectors can achieve a promising performance on only one or two benchmarks. Detectors like SAFE \cite{li2025improving}, achieve an over 95\% accuracy on the previous 3 benchmarks, but fails on the latter 3 with about 50\% accuracy, which demonstrates that it is hard to build a well generalized detector on one training source. \textit{\textbf{(2) GAPL exhibits consistent and promising performance}.} The proposed GAPL achieves a 90.4\% average accuracy across 55 test subsets in 6 benchmarks, with a 3.5\% average accuracy improvement over the previous detectors. Additionally, we reach a 94.9\% average precision and over 90\% mAP in most benchmarks, which further validates our promising detection against generated images. 

\begin{table}[t]
   \centering
    \renewcommand{\arraystretch}{1.1}
    \resizebox{0.48\textwidth}{!}{
    \begin{tabular}{l| c | c c c c c}
     \toprule
        \textbf{Method} & \textbf{Venue} &\textbf{FS} & \textbf{UFD} & \textbf{GI} & \textbf{SB} & \textbf{Mean} \\
        \midrule
        \multicolumn{7}{l}{\textit{General Vision Models}} \\
        \midrule
        ResNet \cite{he2016deep}& CVPR2015 & 73.8 &  79.9 & 66.4 & 73.0 & 73.3\\
        CLIP:ViT \cite{CLIP} & ICML2021 & 78.3 & 74.8 & 73.8  & 65.8 & 73.2  \\
        ConvNext \cite{liu2022convnet} & CVPR2022 & \underline{94.5} & 93.7 & 85.7 & 70.8 & 86.2 \\
        Swin-T \cite{liu2021Swin} & ICCV2021 & 93.3 & \underline{95.6} & 89.8 & 80.2 & \underline{89.7} \\
        \midrule
        \multicolumn{7}{l}{\textit{AIGI Detectors}}  \\
        \midrule
        Co-SPY \cite{cheng2025co} & CVPR2025 & 49.4 & 50.0 & 51.9 & 50.0 & 50.4 \\
        PatchForen \cite{patchforensics} & ECCV2020 & 54.4 & 55.0 & 50.7 & 60.6 & 55.2 \\
        NPR \cite{tan2024rethinking} & CVPR2024 & 51.1 & 57.5 & 56.4 & 56.5 & 55.4 \\
        UniFD \cite{ojha2023towards} & CVPR2023 & 62.4 & 50.1 & 63.6 & 63.3 & 59.8  \\
        D3 \cite{yang2025d3} & CVPR2025 & 75.0 & 57.0& 59.2 & 72.2 & 65.9 \\
        Effort \cite{yan2024orthogonal} & ICML2025 & 84.4 & 66.3 & 71.4 & \underline{88.0} & 77.5 \\ 
        AIDE \cite{yan2024sanity} & ICLR2025 & 86.8 & 90.5 & \underline{96.6} & 67.0 & 85.2 \\
        \midrule
        \rowcolor{cvprblue!20}
        GAPL (Ours) & - & \textbf{97.2} & \textbf{97.2} & \textbf{96.7} & \textbf{91.1} & \textbf{95.5} \\
        \bottomrule
    \end{tabular}
    }
    \caption{Comparison on different model architecture. We re-implement these detection method on exactly the same data as ours, including prototype set and scaling up set and report their binary accuracy and mean accuracy on 4 benchmarks. }
    \label{tab:exp2}
\end{table}

\subsection{Cross-Model Comparison in scaling up setting (RQ.2)}
 To ensure fair comparison and evaluate the inherent ability of the proposed architecture, we conduct an experiment where all methods are trained on the same dataset with ours. Apart from AIGI detectors, we also select some general vision models, including CNN-based \cite{he2016deep, liu2022convnet} and Transformer-based \cite{CLIP, liu2021Swin} as baselines. The results in Tab.~\ref{tab:exp2} reveal the following:
 \textit{\textbf{(1) GAPL has the best mean performance in scaling up setting}.} Among the all detectors, GAPL was the best in terms of accuracy. This indicates our design, integrating priors of pretrained encoder and forgery-related knowledge, is beneficial in scaling up setting. \textit{\textbf{(2) Common AIGI detectors perform even worse when data scales}.} Previous AIGI detectors not only perform worse than general vision models, but also worse than the counterpart that trained on single source. This indicates that previous AIGI detectors are often tailored to specific artifact (e.g VAE decode artifact), rendering them ineffective at scale where diver generative model dilute these priors. 

 \begin{table}[t]
   \centering
    \resizebox{0.42\textwidth}{!}{
    \begin{tabular}{c| c c c | c c}
    \toprule
     \multirow{2}{*}{Group} & \multicolumn{3}{c|}{\textbf{Module}} & \multicolumn{2}{c}{\textbf{Metrics}} \\
        \cmidrule{2-6}
       & \multicolumn{1}{c|}{PCA} & \multicolumn{1}{c|}{PM} & \multicolumn{1}{c|}{LoRA} & MAcc & MAp \\
     \midrule
       1  & \ding{55} & \ding{55} & \ding{55} & 60.05 & 66.07 \\
       2  & \ding{55} & \ding{51} & \ding{55} & 68.59 & 72.43 \\
       3  & \ding{55} & \ding{55} & \ding{51} & 88.52 & 97.91 \\
       4  & \ding{51} & \ding{51} & \ding{55} & 71.88 & 82.18 \\
       5  & \ding{55} & \ding{51} & \ding{51} & 90.35 & 95.40 \\
       \midrule
       Ours & \ding{51} & \ding{51} & \ding{51} & 95.54 & 98.97 \\
       \bottomrule
    \end{tabular}
        } 
        \caption{Ablation study regarding the designed modules. All the results reported are mean of the 4 benchmarks.}
    \label{tab:ablation}
\end{table}
   
\subsection{Effect of the designed modules (RQ.3)}
    To evaluate the impact of the designed modules, we conduct ablation studies on the three components in the proposed GAPL: (PCA) creating extracting prototype from generator via PCA, prototype matching via cross attention (PM) and LoRA finetuning. The results are shown in Tab.~\ref{tab:ablation}. From the results, we can infer that:
   \textit{\textbf{(1) PCA proposed in STAGE I effectively extracts generator-aware concepts}.} Compare performance in group 1,2 and 4, prototype mapping gain an accuracy performance by 8.54\% while PCA further improves 3.28\% by injecting concepts related to generators.
    \textit{\textbf{(2) PM and LoRA work cooperatively to effectively inject large scale generative knowledge to pretrained encoder}.} By comparing the performance of group 2,3 and 5, we observe a gradual performance improvement. This indicates that while the PM module provides the core structure for prototype matching, LoRA finetuning is also essential to adapt encoder to this prototype space, enabling it to effectively align with the extracted generator-aware prototypes.

\subsection{Exploration of selecting Prototype Set (RQ.4)} 

\begin{figure}[t]
    \centering
    \includegraphics[width=\linewidth]{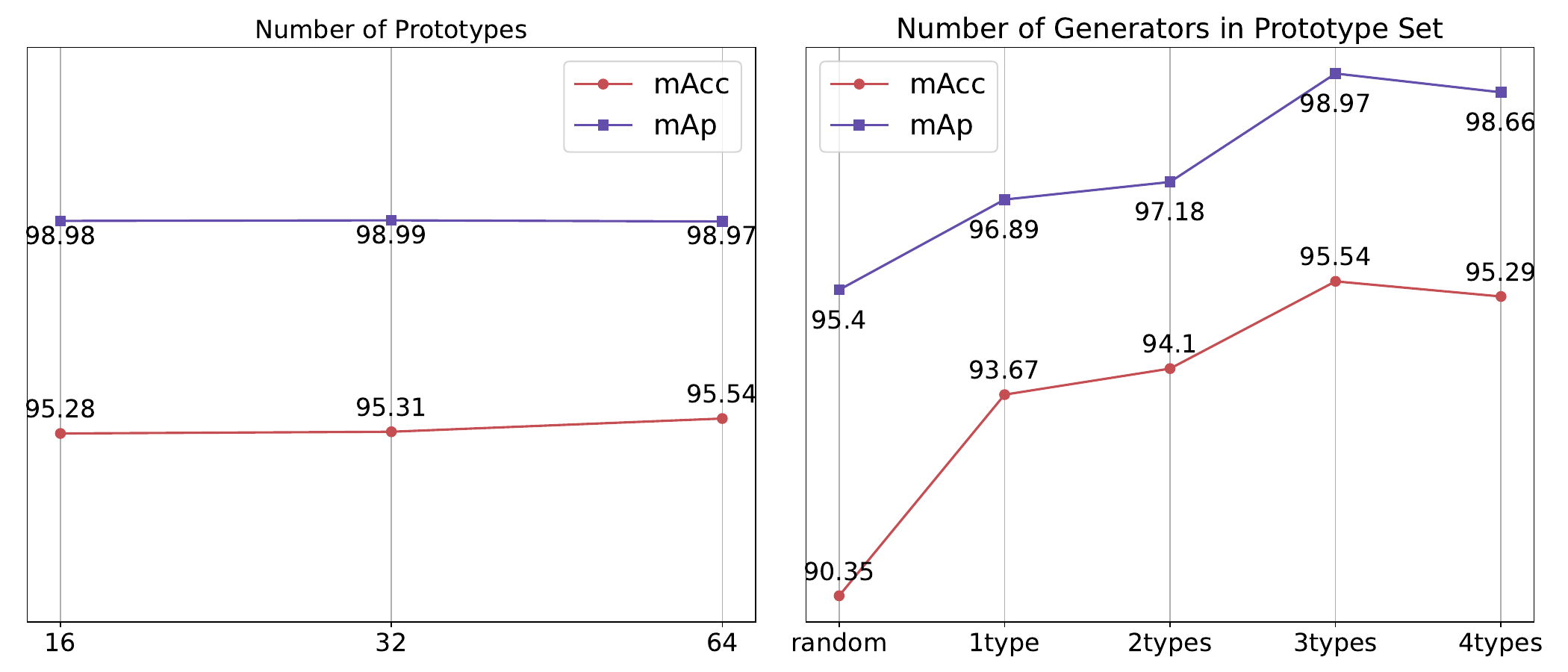}
    \caption{Ablation study regarding the strategy of extracting prototypes. We compare number of prototypes retain in PCA and types of generators in Prototype Set. }
    \label{fig:hyperparam}
\end{figure}

    To explore the influence of the number of prototypes retain in PCA and the selection of prototype set, we conduct experiment in multiple group of  settings. The results are presented in Fig.~\ref{fig:hyperparam}. Results reveal that: \textit{\textbf{(1) Number of prototypes slightly affect performance.}} Model performance slightly improves as $N$ increases, starting from 16 prototypes with a 95.28\% accuracy, keep doubling the number improves it by 0.03\% and 0.23\%. \textit{\textbf{(2) More generator to build prototypes yield performance gains.}} Randomly generating some prototype vector has a performance of 90.35\%, while adding more generator gains performance by 3.22\%, 0.42\% and 1.44\%. But as we try to add generators, the performance goes down, this indicates that the ``a few'' in \textit{\textbf{turn thousand into a few}} should not be too many, three or four is abundant for the model to adjust.

\subsection{Evaluation on robustness (RQ.5)}

\begin{figure}[!t]
    \centering
    \includegraphics[width=\linewidth]{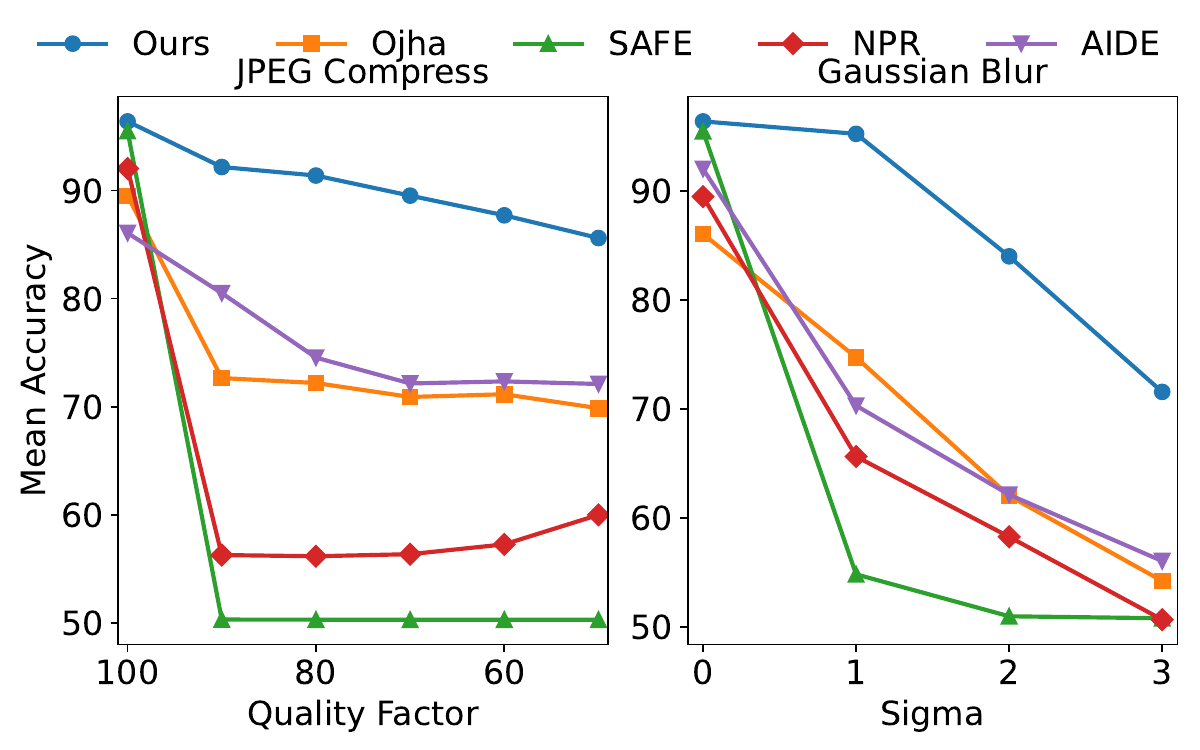}
    \caption{
            Comparison of \textbf{Robustness performance} regarding post process on GenImage dataset. Our model achieve the best overall performance on post process with the minimum degradation.}
        \label{fig:robustness}
\end{figure}

    We evaluate robustness under two common post-processing operations, JPEG compression and Gaussian blur. We compare GAPL with representative AIGI detectors on GenImage, using checkpoints trained on the SD1.4 subset to ensure a comparable starting point. Fig.~\ref{fig:robustness} shows that: \textit{\textbf{(1) GAPL remains the most robust.}} It achieves the best overall performance, with only 11.09\% and 25.12\% degradation under extreme JPEG compression and Gaussian blur, respectively. \textit{\textbf{(2) Frequency-based detectors are highly fragile to post-processing.}} Methods such as AIDE and SAFE degrade severely under JPEG compression, with even mild compression reducing performance to near chance level, indicating limited reliability in real-world settings.

\subsection{Visualization on learned prototypes} 
\begin{figure}[!t]
        \centering
        \includegraphics[width=\linewidth]{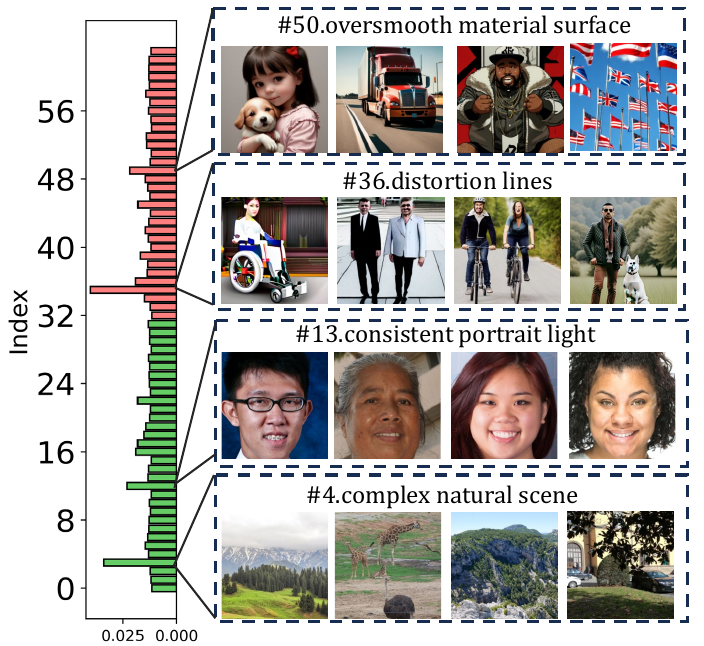} 
        \caption{
            \textbf{Average attention scores} on the validation set. Column in red and green means prototypes extracted with real images and generated images, respectively. We collected images that exhibit high attention score for a specific prototype and discovered that they have visual features in common.
        }
        \label{fig:visualization} 
\end{figure}

    To validate \textbf{how our model utilizes prototypes}, we calculate the average attention scores between prototypes and images on the validation set, we plot the results in Fig.~\ref{fig:visualization}. From the figure, \textit{\textbf{(1) certain prototypes are consistently utilized by the model}.} This indicates that some pattern were common in generated images and real images. Subsequently, to uncover what hide behind these frequently utilized prototypes, we clustered the images that exhibited particularly high attention scores for a specific prototype. We find that \textit{\textbf{(2) images pay extreme attention to a certain prototype share some visual characteristics}.} In generated images, those feature distorted objects and unrealistic lines pay high attention to \textit{Prototype \#36}, images pay attention to \textit{Prototype \# 50} usually contain over smooth material surface. As for real images, complex natural scene and consistent portrait light make them look real, which frequently exist in clusters of \textit{Prototype \#4} and \textit{Prototype \#13}. This finding indicates that our method indeed learned some sort of pattern and the mapping between image with these prototypes.

\section{Conclusion}
    In this paper, we study AI-generated image detection in the scaling-up regime. We identify the key challenges of scaling detector training with diverse generators and highlight the limitations of existing methods. To address these issues, we propose GAPL, which is designed to better exploit data from a wide range of generators. Extensive experiments show that scaling up training is essential for building detectors that generalize across diverse generative models, and that GAPL delivers consistent performance under this setting, taking a step toward universal AIGI detection.

    \textbf{Limitation.} Although our method achieves strong and consistent performance on existing benchmarks, its behavior on truly unseen domains remains unclear, especially for future generative models built upon fundamentally new techniques. Whether such models will still leave detectable artifacts, or become increasingly difficult to distinguish from real images, remains an open problem for future research.

\section*{Acknowledgments}
{\raggedright
This work was supported in part by the National Natural Science Foundation of China under Grant 72434005, Grant 72225011 , Grant 72293575, Grant U24B20179 and Grant 62336001.
\par}
{
    \small
    \bibliographystyle{ieeenat_fullname}
    \bibliography{main}

@String(CVPR= {IEEE Conf. Comput. Vis. Pattern Recog.})

@String(ICCV= {Int. Conf. Comput. Vis.})

@String(ICLR = {Int. Conf. Learn. Represent.})

@String(AAAI = {AAAI})

@String(CVPR  = {CVPR})

@String(ICCV  = {ICCV})

@String(ICLR  = {ICLR})

@inproceedings{wang2020cnn,
  title={CNN-generated images are surprisingly easy to spot... for now},
  author={Wang, Sheng-Yu and Wang, Oliver and Zhang, Richard and Owens, Andrew and Efros, Alexei A},
  booktitle={Proceedings of the IEEE/CVF conference on computer vision and pattern recognition},
  pages={8695--8704},
  year={2020}
}

@article{yan2024sanity,
  title={A sanity check for ai-generated image detection},
  author={Yan, Shilin and Li, Ouxiang and Cai, Jiayin and Hao, Yanbin and Jiang, Xiaolong and Hu, Yao and Xie, Weidi},
  journal={arXiv preprint arXiv:2406.19435},
  year={2024}
}

@inproceedings{ojha2023towards,
  title={Towards universal fake image detectors that generalize across generative models},
  author={Ojha, Utkarsh and Li, Yuheng and Lee, Yong Jae},
  booktitle={Proceedings of the IEEE/CVF Conference on Computer Vision and Pattern Recognition},
  pages={24480--24489},
  year={2023}
}

@inproceedings{tan2024rethinking,
  title={Rethinking the up-sampling operations in cnn-based generative network for generalizable deepfake detection},
  author={Tan, Chuangchuang and Zhao, Yao and Wei, Shikui and Gu, Guanghua and Liu, Ping and Wei, Yunchao},
  booktitle={Proceedings of the IEEE/CVF Conference on Computer Vision and Pattern Recognition},
  pages={28130--28139},
  year={2024}
}

@inproceedings{tan2024frequency,
  title={Frequency-aware deepfake detection: Improving generalizability through frequency space domain learning},
  author={Tan, Chuangchuang and Zhao, Yao and Wei, Shikui and Gu, Guanghua and Liu, Ping and Wei, Yunchao},
  booktitle={Proceedings of the AAAI Conference on Artificial Intelligence},
  volume={38},
  
  pages={5052--5060},
  year={2024}
}

@inproceedings{liu2024forgery,
  title={Forgery-aware adaptive transformer for generalizable synthetic image detection},
  author={Liu, Huan and Tan, Zichang and Tan, Chuangchuang and Wei, Yunchao and Wang, Jingdong and Zhao, Yao},
  booktitle={Proceedings of the IEEE/CVF Conference on Computer Vision and Pattern Recognition},
  pages={10770--10780},
  year={2024}
}

@inproceedings{chen2024drct,
  title={Drct: Diffusion reconstruction contrastive training towards universal detection of diffusion generated images},
  author={Chen, Baoying and Zeng, Jishen and Yang, Jianquan and Yang, Rui},
  booktitle={Forty-first International Conference on Machine Learning},
  year={2024}
}

@article{yan2024orthogonal,
  title={Orthogonal Subspace Decomposition for Generalizable AI-Generated Image Detection},
  author={Yan, Zhiyuan and Wang, Jiangming and Jin, Peng and Zhang, Ke-Yue and Liu, Chengchun and Chen, Shen and Yao, Taiping and Ding, Shouhong and Wu, Baoyuan and Yuan, Li},
  journal={arXiv preprint arXiv:2411.15633},
  year={2024}
}

@inproceedings{tan2025c2p,
  title={C2p-clip: Injecting category common prompt in clip to enhance generalization in deepfake detection},
  author={Tan, Chuangchuang and Tao, Renshuai and Liu, Huan and Gu, Guanghua and Wu, Baoyuan and Zhao, Yao and Wei, Yunchao},
  booktitle={Proceedings of the AAAI Conference on Artificial Intelligence},
  volume={39},
  
  pages={7184--7192},
  year={2025}
}

@inproceedings{CLIP,
  title={Learning transferable visual models from natural language supervision},
  author={Radford, Alec and Kim, Jong Wook and Hallacy, Chris and Ramesh, Aditya and Goh, Gabriel and Agarwal, Sandhini and Sastry, Girish and Askell, Amanda and Mishkin, Pamela and Clark, Jack and others},
  booktitle={International conference on machine learning},
  pages={8748--8763},
  year={2021},
  organization={PmLR}
}

@inproceedings{he2016deep,
  title={Deep residual learning for image recognition},
  author={He, Kaiming and Zhang, Xiangyu and Ren, Shaoqing and Sun, Jian},
  booktitle={Proceedings of the IEEE conference on computer vision and pattern recognition},
  pages={770--778},
  year={2016}
}

@inproceedings{cheng2025co,
  title={CO-SPY: Combining Semantic and Pixel Features to Detect Synthetic Images by AI},
  author={Cheng, Siyuan and Lyu, Lingjuan and Wang, Zhenting and Zhang, Xiangyu and Sehwag, Vikash},
  booktitle={Proceedings of the Computer Vision and Pattern Recognition Conference},
  pages={13455--13465},
  year={2025}
}

@inproceedings{
rajan2025aligned,
title={Aligned Datasets Improve Detection of Latent Diffusion-Generated Images},
author={Anirudh Sundara Rajan and Utkarsh Ojha and Jedidiah Schloesser and Yong Jae Lee},
booktitle={The Thirteenth International Conference on Learning Representations},
year={2025},
url={https://openreview.net/forum?id=doBkiqESYq}
}

@inproceedings{Guillaro2024biasfree,
   author    = {Guillaro, Fabrizio and Zingarini, Giada and Usman, Ben and Sud, Avneesh and Cozzolino, Davide and Verdoliva, Luisa},
   title     = {A Bias-Free Training Paradigm for More General AI-generated Image Detection},
   booktitle = {Proceedings of the Computer Vision and Pattern Recognition Conference (CVPR)},
   month     = {June},
   year      = {2025},
   pages     = {18685-18694}
}

@inproceedings{chen2025dda,
  title={Dual Data Alignment Makes AI-Generated Image Detector Easier Generalizable},
  author={Chen, Ruoxin and Xi, Junwei and Yan, Zhiyuan and Zhang, Ke-Yue and Wu, Shuang and Xie, Jingyi and Chen, Xu and Xu, Lei and Guan, Isabel and Yao, Taiping and Ding, Shouhong},
  booktitle={Advances in Neural Information Processing Systems},
  year={2025}
}

@inproceedings{yang2025d3,
  title={D3: Scaling Up Deepfake Detection by Learning from Discrepancy},
  author={Yang, Yongqi and Qian, Zhihao and Zhu, Ye and Russakovsky, Olga and Wu, Yu},
  booktitle={Proceedings of the IEEE/CVF Conference on Computer Vision and Pattern Recognition},
  year={2025}
}

@inproceedings{park2025commfor,
    author    = {Park, Jeongsoo and Owens, Andrew},
    title     = {Community Forensics: Using Thousands of Generators to Train Fake Image Detectors},
    booktitle = {Proceedings of the Computer Vision and Pattern Recognition Conference (CVPR)},
    month     = {June},
    year      = {2025},
    pages     = {8245-8257}
}

@inproceedings{li2025improving,
  title={Improving synthetic image detection towards generalization: An image transformation perspective},
  author={Li, Ouxiang and Cai, Jiayin and Hao, Yanbin and Jiang, Xiaolong and Hu, Yao and Feng, Fuli},
  booktitle={Proceedings of the 31st ACM SIGKDD Conference on Knowledge Discovery and Data Mining V. 1},
  pages={2405--2414},
  year={2025}
}

@inproceedings{vaswani2017attention,
 author = {Vaswani, Ashish and Shazeer, Noam and Parmar, Niki and Uszkoreit, Jakob and Jones, Llion and Gomez, Aidan N and Kaiser, \L ukasz and Polosukhin, Illia},
 booktitle = {Advances in Neural Information Processing Systems},
 editor = {I. Guyon and U. Von Luxburg and S. Bengio and H. Wallach and R. Fergus and S. Vishwanathan and R. Garnett},
 pages = {},
 publisher = {Curran Associates, Inc.},
 title = {Attention is All you Need},
 url = {https://proceedings.neurips.cc/paper_files/paper/2017/file/3f5ee243547dee91fbd053c1c4a845aa-Paper.pdf},
 volume = {30},
 year = {2017}
}

@Article{liu2022convnet,
  author  = {Zhuang Liu and Hanzi Mao and Chao-Yuan Wu and Christoph Feichtenhofer and Trevor Darrell and Saining Xie},
  title   = {A ConvNet for the 2020s},
  journal = {Proceedings of the IEEE/CVF Conference on Computer Vision and Pattern Recognition (CVPR)},
  year    = {2022},
}

@inproceedings{liu2021Swin,
  title={Swin Transformer: Hierarchical Vision Transformer using Shifted Windows},
  author={Liu, Ze and Lin, Yutong and Cao, Yue and Hu, Han and Wei, Yixuan and Zhang, Zheng and Lin, Stephen and Guo, Baining},
  booktitle={Proceedings of the IEEE/CVF International Conference on Computer Vision (ICCV)},
  year={2021}
}

@inproceedings{karras2018progressive,
  title={Progressive Growing of GANs for Improved Quality, Stability, and Variation},
  author={Karras, Tero and others},
  booktitle={International Conference on Learning Representations},
  year={2018}
}

@inproceedings{karras2019style,
  title={A style-based generator architecture for generative adversarial networks},
  author={Karras, Tero and others},
  booktitle={Proceedings of the IEEE/CVF Conference on Computer Vision and Pattern Recognition},
  pages={4401--4410},
  year={2019}
}

@inproceedings{BigGAN,
  title={Large Scale GAN Training for High Fidelity Natural Image Synthesis},
  author={Brock, Andrew and others},
  booktitle={International Conference on Learning Representations},
  year={2018}
}

@inproceedings{CycleGAN,
  title={Unpaired image-to-image translation using cycle-consistent adversarial networks},
  author={Zhu, Jun-Yan and others},
  booktitle={Proceedings of the IEEE International Conference on Computer Vision},
  pages={2223--2232},
  year={2017}
}

@inproceedings{choi2018stargan,
  title={Stargan: Unified generative adversarial networks for multi-domain image-to-image translation},
  author={Choi, Yunjey and others},
  booktitle={Proceedings of the IEEE Conference on Computer Vision and Pattern Recognition},
  pages={8789--8797},
  year={2018}
}

@inproceedings{GauGAN,
  title={Semantic image synthesis with spatially-adaptive normalization},
  author={Park, Taesung and others},
  booktitle={Proceedings of the IEEE/CVF Conference on Computer Vision and Pattern Recognition},
  pages={2337--2346},
  year={2019}
}

@inproceedings{Deepfake,
  title={Faceforensics++: Learning to detect manipulated facial images},
  author={Rossler, Andreas and others},
  booktitle={Proceedings of the IEEE/CVF International Conference on Computer Vision},
  pages={1--11},
  year={2019}
}

@article{dhariwal2021diffusion,
  title={Diffusion models beat gans on image synthesis},
  author={Dhariwal, Prafulla and others},
  journal={Advances in Neural Information Processing Systems},
  volume={34},
  pages={8780--8794},
  year={2021}
}

@article{nichol2021glide,
  title={Glide: Towards photorealistic image generation and editing with text-guided diffusion models},
  author={Nichol, Alex and Dhariwal, Prafulla and Ramesh, Aditya and Shyam, Pranav and Mishkin, Pamela and McGrew, Bob and Sutskever, Ilya and Chen, Mark},
  journal={arXiv preprint arXiv:2112.10741},
  year={2021}
}

@article{zhu2024genimage,
  title={Genimage: A million-scale benchmark for detecting ai-generated image},
  author={Zhu, Mingjian and Chen, Hanting and Yan, Qiangyu and Huang, Xudong and Lin, Guanyu and Li, Wei and Tu, Zhijun and Hu, Hailin and Hu, Jie and Wang, Yunhe},
  journal={Advances in Neural Information Processing Systems},
  volume={36},
  year={2024}
}

@Misc{Midjourney,
  title =        {Midjourney},
  author =       {},
  howpublished = {\url{In https://www.midjourney.com/home/}},
  year =         {2022}
}

@Misc{Wukong,
  title =        {Wukong, 2022. 5},
  author =       {},
  howpublished = {\url{In https://xihe.mindspore.cn/modelzoo/wukong}},
  year =         {2022. 5}
}

@misc{AdobeFirefly,
  author       = {{Adobe}},
  title        = {Adobe Firefly},
  howpublished = {\url{https://firefly.adobe.com/}},
  note         = {Accessed: 2025-11-04},
  year         = {2025}
}

@inproceedings{gu2022vector,
  title={Vector quantized diffusion model for text-to-image synthesis},
  author={Gu, Shuyang and Chen, Dong and Bao, Jianmin and Wen, Fang and Zhang, Bo and Chen, Dongdong and Yuan, Lu and Guo, Baining},
  booktitle={Proceedings of the IEEE/CVF Conference on Computer Vision and Pattern Recognition},
  pages={10696--10706},
  year={2022}
}

@inproceedings{
hu2022lora,
title={Lo{RA}: Low-Rank Adaptation of Large Language Models},
author={Edward J Hu and Yelong Shen and Phillip Wallis and Zeyuan Allen-Zhu and Yuanzhi Li and Shean Wang and Lu Wang and Weizhu Chen},
booktitle={International Conference on Learning Representations},
year={2022},
url={https://openreview.net/forum?id=nZeVKeeFYf9}
}

@inproceedings{zhang2025towards,
  title={Towards Universal AI-Generated Image Detection by Variational Information Bottleneck Network},
  author={Zhang, Haifeng and He, Qinghui and Bi, Xiuli and Li, Weisheng and Liu, Bo and Xiao, Bin},
  booktitle={Proceedings of the Computer Vision and Pattern Recognition Conference},
  pages={23828--23837},
  year={2025}
}

@article{wen2025spot,
  title={Spot the fake: Large multimodal model-based synthetic image detection with artifact explanation},
  author={Wen, Siwei and Ye, Junyan and Feng, Peilin and Kang, Hengrui and Wen, Zichen and Chen, Yize and Wu, Jiang and Wu, Wenjun and He, Conghui and Li, Weijia},
  journal={arXiv preprint arXiv:2503.14905},
  year={2025}
}

@article{synthbuster,
  author={Bammey, Quentin},
  journal={IEEE Open Journal of Signal Processing}, 
  title={Synthbuster: Towards Detection of Diffusion Model Generated Images}, 
  year={2024},
  volume={5},
  number={},
  pages={1-9},
  doi={10.1109/OJSP.2023.3337714}}

@inproceedings{rombach2022high,
    author    = {Rombach, Robin and Blattmann, Andreas and Lorenz, Dominik and Esser, Patrick and Ommer, Bj\"orn},
    title     = {High-Resolution Image Synthesis With Latent Diffusion Models},
    booktitle = {Proceedings of the IEEE/CVF Conference on Computer Vision and Pattern Recognition (CVPR)},
    month     = {June},
    year      = {2022},
    pages     = {10684-10695}
}

@misc{liang2025ferretnet,
      title={FerretNet: Efficient Synthetic Image Detection via Local Pixel Dependencies}, 
      author={Shuqiao Liang and Jian Liu and Renzhang Chen and Quanlong Guan},
      year={2025},
      eprint={2509.20890},
      archivePrefix={arXiv},
      primaryClass={cs.CV},
      url={https://arxiv.org/abs/2509.20890}, 
}

@article{fisher1936use,
  author    = {Fisher, R. A.},
  title     = {The Use of Multiple Measurements in Taxonomic Problems},
  journal   = {Annals of Eugenics},
  year      = {1936},
  volume    = {7},
  number    = {2},
  pages     = {179--188},
  doi       = {10.1111/j.1469-1809.1936.tb02137.x}
}

@inproceedings{imagenet,
  author={Deng, Jia and Dong, Wei and Socher, Richard and Li, Li-Jia and Kai Li and Li Fei-Fei},
  booktitle={2009 IEEE Conference on Computer Vision and Pattern Recognition}, 
  title={ImageNet: A large-scale hierarchical image database}, 
  year={2009},
  volume={},
  number={},
  pages={248-255},
  doi={10.1109/CVPR.2009.5206848}}

@inproceedings{bihpf,
  author={Jeong, Yonghyun and Kim, Doyeon and Min, Seungjai and Joe, Seongho and Gwon, Youngjune and Choi, Jongwon},
  booktitle={2022 IEEE/CVF Winter Conference on Applications of Computer Vision (WACV)}, 
  title={BiHPF: Bilateral High-Pass Filters for Robust Deepfake Detection}, 
  year={2022},
  volume={},
  number={},
  pages={2878-2887},
  doi={10.1109/WACV51458.2022.00293}}

@article{red116,
  author={Asnani, Vishal and Yin, Xi and Hassner, Tal and Liu, Xiaoming},
  journal={IEEE Transactions on Pattern Analysis and Machine Intelligence}, 
  title={Reverse Engineering of Generative Models: Inferring Model Hyperparameters From Generated Images}, 
  year={2023},
  volume={45},
  number={12},
  pages={15477-15493},
  doi={10.1109/TPAMI.2023.3301451}}

@inproceedings{red140,
 author = {Guo, Xiao and Asnani, Vishal and Liu, Sijia and Liu, Xiaoming},
 booktitle = {Advances in Neural Information Processing Systems},
 doi = {10.52202/079017-3711},
 editor = {A. Globerson and L. Mackey and D. Belgrave and A. Fan and U. Paquet and J. Tomczak and C. Zhang},
 pages = {116899--116932},
 publisher = {Curran Associates, Inc.},
 title = {Tracing Hyperparameter Dependencies for Model Parsing via Learnable Graph Pooling Network},
 url = {https://proceedings.neurips.cc/paper_files/paper/2024/file/d412468559da4e2ac41a0285b3a079cd-Paper-Conference.pdf},
 volume = {37},
 year = {2024}
}

@article{wang2023dire,
  title={DIRE for Diffusion-Generated Image Detection},
  author={Wang, Zhendong and Bao, Jianmin and Zhou, Wengang and Wang, Weilun and Hu, Hezhen and Chen, Hong and Li, Houqiang},
  journal={arXiv preprint arXiv:2303.09295},
  year={2023}
}

@inproceedings{luo2024lare,
    author    = {Luo, Yunpeng and Du, Junlong and Yan, Ke and Ding, Shouhong},
    title     = {LaRE{\textasciicircum}2: Latent Reconstruction Error Based Method for Diffusion-Generated Image Detection},
    booktitle = {Proceedings of the IEEE/CVF Conference on Computer Vision and Pattern Recognition (CVPR)},
    month     = {June},
    year      = {2024},
    pages     = {17006-17015}
}

@misc{wu2025qwenimage,
      title={Qwen-Image Technical Report}, 
      author={Chenfei Wu and Jiahao Li and Jingren Zhou and Junyang Lin and Kaiyuan Gao and Kun Yan and Sheng-ming Yin and Shuai Bai and Xiao Xu and Yilei Chen and Yuxiang Chen and Zecheng Tang and Zekai Zhang and Zhengyi Wang and An Yang and Bowen Yu and Chen Cheng and Dayiheng Liu and Deqing Li and Hang Zhang and Hao Meng and Hu Wei and Jingyuan Ni and Kai Chen and Kuan Cao and Liang Peng and Lin Qu and Minggang Wu and Peng Wang and Shuting Yu and Tingkun Wen and Wensen Feng and Xiaoxiao Xu and Yi Wang and Yichang Zhang and Yongqiang Zhu and Yujia Wu and Yuxuan Cai and Zenan Liu},
      year={2025},
      eprint={2508.02324},
      archivePrefix={arXiv},
      primaryClass={cs.CV},
      url={https://arxiv.org/abs/2508.02324}, 
}

@article{loshchilov2017decoupled,
  title   = {Decoupled Weight Decay Regularization},
  author  = {Loshchilov, Ilya and Hutter, Frank},
  journal = {arXiv preprint arXiv:1711.05101},
  year    = {2017},
  note    = {Submitted November 14, 2017; revised January 4, 2019},
  url     = {https://arxiv.org/abs/1711.05101}
}

@article{dosovitskiy2020vit,
  title={An Image is Worth 16x16 Words: Transformers for Image Recognition at Scale},
  author={Dosovitskiy, Alexey and Beyer, Lucas and Kolesnikov, Alexander and Weissenborn, Dirk and Zhai, Xiaohua and Unterthiner, Thomas and  Dehghani, Mostafa and Minderer, Matthias and Heigold, Georg and Gelly, Sylvain and Uszkoreit, Jakob and Houlsby, Neil},
  journal={ICLR},
  year={2021}
}

@misc{DeciFoundationModels,
title = {DeciDiffusion 2.0},
author = {DeciAI Research Team},
year = {2024},
url={\url{https://huggingface.co/deci/decidiffusion-v2-0}},
}

@inproceedings{tao2023galip,
  title={Galip: Generative adversarial clips for text-to-image synthesis},
  author={Tao, Ming and Bao, Bing-Kun and Tang, Hao and Xu, Changsheng},
  booktitle={Proceedings of the IEEE/CVF Conference on Computer Vision and Pattern Recognition},
  pages={14214--14223},
  year={2023}
}

@misc{kandinsky2023models,
  title={Kandinsky 2.2},
  author={Shakhmatov, Arseniy and Razzhigaev, Anton and Nikolich, Aleksandr and Arkhipkin, Vladimir and Pavlov, Igor and Kuznetsov, Andrey and Dimitrov, Denis},
  year={2023},
  howpublished = {\url{https://github.com/ai-forever/Kandinsky-2}}
}

@mist{KvikontentMidjourney,
  title={Kvikontent-Midjourney V6},
  author={Kvikontent},
  year={2023},
  howpublished={https://huggingface.co/Kvikontent/midjourney-v6}
}

@inproceedings{pernias2023stablecascade,
  title={W{\"u}rstchen: An Efficient Architecture for Large-Scale Text-to-Image Diffusion Models},
  author={Pernias, Pablo and Rampas, Dominic and Richter, Mats Leon and Pal, Christopher and Aubreville, Marc},
  booktitle={The Twelfth International Conference on Learning Representations},
  year={2023}
}

@article{luo2023lcmLora,
  title={Lcm-lora: A universal stable-diffusion acceleration module},
  author={Luo, Simian and Tan, Yiqin and Patil, Suraj and Gu, Daniel and von Platen, Patrick and Passos, Apolin{\'a}rio and Huang, Longbo and Li, Jian and Zhao, Hang},
  journal={arXiv preprint arXiv:2311.05556},
  year={2023}
}

@inproceedings{tao2022df,
  title={DF-gan: A simple and effective baseline for text-to-image synthesis},
  author={Tao, Ming and Tang, Hao and Wu, Fei and Jing, Xiao-Yuan and Bao, Bing-Kun and Xu, Changsheng},
  booktitle={Proceedings of the IEEE/CVF conference on computer vision and pattern recognition},
  pages={16515--16525},
  year={2022}
}

@inproceedings{crowson2024scalable,
    title = 	 {Scalable High-Resolution Pixel-Space Image Synthesis with Hourglass Diffusion Transformers},
    author =       {Crowson, Katherine and Baumann, Stefan Andreas and Birch, Alex and Abraham, Tanishq Mathew and Kaplan, Daniel Z and Shippole, Enrico},
    booktitle = 	 {Proceedings of the 41st International Conference on Machine Learning},
    pages = 	 {9550--9575},
    year = 	 {2024},
    editor = 	 {Salakhutdinov, Ruslan and Kolter, Zico and Heller, Katherine and Weller, Adrian and Oliver, Nuria and Scarlett, Jonathan and Berkenkamp, Felix},
    volume = 	 {235},
    series = 	 {Proceedings of Machine Learning Research},
    month = 	 {21--27 Jul},
    publisher =    {PMLR},
    url = 	 {https://proceedings.mlr.press/v235/crowson24a.html},
}

@misc{blackforestlabs2024flux,
  title = {{FLUX.1}: Speeding up Text-to-Image Generation},
  author = {{Black Forest Labs}},
  year = {2025},
  publisher = {Black Forest Labs},
  howpublished = {\url{https://blackforestlabs.ai}},
  note = {Accessed: 2025-11-26}
}

@misc{midjourneyv6,
  title = {Midjourney v6},
  author = {{Midjourney, Inc.}},
  year = {2025},
  howpublished = {\url{https://www.midjourney.com}},
  note = {AI model version 6.0, Accessed: 2025-11-26}
}

@misc{banana,
  title = {Nano Banana},
  author = {{Google, Inc.}},
  year = {2025},
  howpublished = {\url{https://www.nano-banana.com/}},
  note = {Accessed: 2025-08-29}
}

@inproceedings{tanreason,
  title={Reason-RFT: Reinforcement Fine-Tuning for Visual Reasoning of Vision Language Models},
  author={Tan, Huajie and Ji, Yuheng and Hao, Xiaoshuai and Chen, Xiansheng and Wang, Pengwei and Wang, Zhongyuan and Zhang, Shanghang},
  booktitle={The Thirty-ninth Annual Conference on Neural Information Processing Systems}
}

@article{ji2025mathsticks,
  title={Mathsticks: A benchmark for visual symbolic compositional reasoning with matchstick puzzles},
  author={Ji, Yuheng and Tan, Huajie and Chi, Cheng and Xu, Yijie and Zhao, Yuting and Zhou, Enshen and Lyu, Huaihai and Wang, Pengwei and Wang, Zhongyuan and Zhang, Shanghang and others},
  journal={arXiv preprint arXiv:2510.00483},
  year={2025}
}

@article{ji2025visualtrans,
  title={Visualtrans: A benchmark for real-world visual transformation reasoning},
  author={Ji, Yuheng and Wang, Yipu and Liu, Yuyang and Hao, Xiaoshuai and Liu, Yue and Zhao, Yuting and Lyu, Huaihai and Zheng, Xiaolong},
  journal={arXiv preprint arXiv:2508.04043},
  year={2025}
}

@article{song2025maniplvm,
  title={Maniplvm-r1: Reinforcement learning for reasoning in embodied manipulation with large vision-language models},
  author={Song, Zirui and Ouyang, Guangxian and Li, Mingzhe and Ji, Yuheng and Wang, Chenxi and Xu, Zixiang and Zhang, Zeyu and Zhang, Xiaoqing and Jiang, Qian and Chen, Zhenhao and others},
  journal={arXiv preprint arXiv:2505.16517},
  year={2025}
}

@inproceedings{ji2025robobrain,
  title={Robobrain: A unified brain model for robotic manipulation from abstract to concrete},
  author={Ji, Yuheng and Tan, Huajie and Shi, Jiayu and Hao, Xiaoshuai and Zhang, Yuan and Zhang, Hengyuan and Wang, Pengwei and Zhao, Mengdi and Mu, Yao and An, Pengju and others},
  booktitle={Proceedings of the Computer Vision and Pattern Recognition Conference},
  pages={1724--1734},
  year={2025}
}

@article{team2025robobrain,
  title={Robobrain 2.0 technical report},
  author={Team, BAAI RoboBrain and Cao, Mingyu and Tan, Huajie and Ji, Yuheng and Chen, Xiansheng and Lin, Minglan and Li, Zhiyu and Cao, Zhou and Wang, Pengwei and Zhou, Enshen and others},
  journal={arXiv preprint arXiv:2507.02029},
  year={2025}
}

@inproceedings{ji2025enhancing,
  title={Enhancing adversarial robustness of vision-language models through low-rank adaptation},
  author={Ji, Yuheng and Liu, Yue and Zhang, Zhicheng and Zhang, Zhao and Zhao, Yuting and Hao, Xiaoshuai and Zhou, Gang and Zhang, Xingwei and Zheng, Xiaolong},
  booktitle={Proceedings of the 2025 International Conference on Multimedia Retrieval},
  pages={550--559},
  year={2025}
}

@inproceedings{bai2025alleviating,
  title={Alleviating Performance Disparity in Adversarial Spatiotemporal Graph Learning Under Zero-Inflated Distribution},
  author={Bai, Songran and Ji, Yuheng and Liu, Yue and Zhang, Xingwei and Zheng, Xiaolong and Zeng, Daniel Dajun},
  booktitle={Proceedings of the AAAI Conference on Artificial Intelligence},
  volume={39},
  number={11},
  pages={11436--11444},
  year={2025}
}

@article{bai2025towards,
  title={Towards a unified understanding of robot manipulation: A comprehensive survey},
  author={Bai, Shuanghao and Song, Wenxuan and Chen, Jiayi and Ji, Yuheng and Zhong, Zhide and Yang, Jin and Zhao, Han and Zhou, Wanqi and Zhao, Wei and Li, Zhe and others},
  journal={arXiv preprint arXiv:2510.10903},
  year={2025}
}

@article{lyu2025egoprompt,
  title={Egoprompt: Prompt pool learning for egocentric action recognition},
  author={Lyu, Huaihai and Chen, Chaofan and Ji, Yuheng and Xu, Changsheng},
  journal={arXiv preprint arXiv:2508.03266},
  year={2025}
}

@article{tan2025roboos,
  title={RoboOS-NeXT: A Unified Memory-based Framework for Lifelong, Scalable, and Robust Multi-Robot Collaboration},
  author={Tan, Huajie and Chi, Cheng and Chen, Xiansheng and Ji, Yuheng and Zhao, Zhongxia and Hao, Xiaoshuai and Lyu, Yaoxu and Cao, Mingyu and Zhao, Junkai and Lyu, Huaihai and others},
  journal={arXiv preprint arXiv:2510.26536},
  year={2025}
}

@article{wang2025crosspoint,
  title   = {Towards Cross-View Point Correspondence in Vision-Language Models},
  author  = {Wang, Yipu and Ji, Yuheng and Liu, Yuyang and Zhou, Enshen and Yang, Ziqiang and Tian, Yuxuan and Qin, Ziheng and Liu, Yue and Tan, Huajie and Chi, Cheng and Ma, Zhiyuan and Zeng, Daniel Dajun and Zheng, Xiaolong},
  journal = {arXiv preprint arXiv:2512.04686},
  year= {2025}
}

@inproceedings{patchforensics,
  title={What makes fake images detectable? Understanding properties that generalize},
  author={Chai, Lucy and Bau, David and Lim, Ser-Nam and Isola, Phillip},
  booktitle={European Conference on Computer Vision},
  year={2020}
 }

@article{li2025towards,
  title={Towards Generalizable AI-Generated Image Detection via Image-Adaptive Prompt Learning},
  author={Li, Yiheng and Tan, Zichang and Xu Guoqing and Lei, Zhen and Zhou, Xu and Yang, Yang},
  journal={arXiv preprint arXiv:2508.01603},
  year={2025}
}

@article{liu2025beyond,
  title={Beyond Artifacts: Real-Centric Envelope Modeling for Reliable AI-Generated Image Detection},
  author={Liu, Ruiqi and Han, Yi and Zhang, Zhengbo and Yao, Liwei and Yan, Zhiyuan and Shen, Jialiang and Chen, ZhiJin and Sun, Boyi and Weng, Lubin and Dong, Jing and others},
  journal={arXiv preprint arXiv:2512.20937},
  year={2025}
}

@article{liu2026mirror,
  title={MIRROR: Manifold Ideal Reference ReconstructOR for Generalizable AI-Generated Image Detection},
  author={Liu, Ruiqi and Cui, Manni and Qin, Ziheng and Yan, Zhiyuan and Chen, Ruoxin and Han, Yi and Li, Zhiheng and Chen, Junkai and Chen, ZhiJin and Lin, Kaiqing and others},
  journal={arXiv preprint arXiv:2602.02222},
  year={2026}
}
}

\clearpage
\setcounter{page}{1}
\maketitlesupplementary

\appendix
We organize the supplementary in the following way:
\begin{itemize}
    \item Sec.~\ref{sec:pre-exp}: Detailed information of the introduced series of dataset build in previous section.
    \item Sec.~\ref{sec:benchmark}: Details of the 6 selected benchmarks. 
    \item Sec.~\ref{sec:baseline}: Introduction to the selected baselines.
    \item Sec.~\ref{sec:detailexp}: Detailed results of each test subsets.
    \item Sec.~\ref{sec:impledet}: Implementation details of the proposed method.
    \item Sec.~\ref{sec:additional_analysis}: Additional derivation and analysis.
    \item Sec.~\ref{sec:future}: Future perspectives towards reasoning-driven and embodied forensics.
\end{itemize}

\section{Settings of the toy dataset}
\label{sec:pre-exp}
In Sec.3, we introduce a series of datasets, comprising generated images from different numbers of generators. In this section we will give the constructing procedure of these datasets.

First, there are 8 generators in GenImage \cite{zhu2024genimage} dataset, in each generator subset, there are 1000 types of object mirroring the 1000 categories of ImageNet-1k \cite{imagenet}. We select generator subset composed of $n_g$ generators based on Tab.~\ref{tab:toydataset}. For each of the 1000 categories, we randomly sample $n_s$ images in each  generator subset. To pair real images, we randomly sample $n_s \times n_g$ images from the original ImageNet dataset. $n_s$ is determined via  $1000 \times n_s \times n_g=8000$.

For the last dataset, which consist of thousands of generators, we leverage Community-Forensics training dataset \cite{park2025commfor} for collecting, which is the same as our training dataset. we randomly sample 2 images from each generator in it, which consist of about 9000 generated images. Then we randomly sample 8,000 real images as before to construct the last dataset in the series.

\begin{table}[h]
    \centering
    \begin{tabular}{c|c|c}
    \toprule
     group & $n_g$ & Generator(s) \\ 
     \midrule
        1 &  1  & SDv1.4 \\
        2 &  2  & SDv1.4, BigGAN \\ 
        3 &  4 &  SDv1.4, BigGAN, VQDM, Glide \\
        4 &  8 & All GenImage \\
        \bottomrule
    \end{tabular}
    \caption{Generators used to build our datasets.}
    \label{tab:toydataset}
\end{table}

\section{Benchmarks}
\label{sec:benchmark}

We select 6 benchmarks to represent most existed generative models to evaluate the methods. Though some subsets have same or similar architecture, their training condition, sampling strategy and semantic content are not quite the same. Thus we preserve all subsets that have same name to simulate a variety of generated images.

\vspace{2mm}
\noindent
\textbf{Forensic Synthetic} \cite{wang2020cnn} contains a series of CNN-generated images, we select its GAN variants, including ProGAN \cite{karras2018progressive}, StyleGAN \cite{karras2019style}, StyleGAN2, CycleGAN \cite{CycleGAN}, StarGAN \cite{choi2018stargan}, GauGAN \cite{GauGAN}, BigGAN \cite{BigGAN}, and Deepfake \cite{Deepfake} for forgery face.

\vspace{2mm}
\noindent
\textbf{UFD} \cite{ojha2023towards} datasets expand the dataset above by introducing several early diffusion models and commercial APIs, including latent diffusion model \cite{rombach2022high}, Glide \cite{nichol2021glide} and Guided \cite{dhariwal2021diffusion} diffusion model.

\vspace{2mm}
\noindent
\textbf{GenImage} \cite{zhu2024genimage} provide a dataset trained on ImageNet-1k. It has 8 generative models in both GANs, Diffusion Models and Commercial APIs, including BigGAN \cite{BigGAN}, VQDM \cite{gu2022vector}, Stable Diffusions, Wukong \cite{Wukong}, ADM \cite{dhariwal2021diffusion} and Midjourney \cite{Midjourney}.

\vspace{2mm}
\noindent
\textbf{SynthBuster} \cite{synthbuster} provide an aligned dataset, where real images and generated images are all in PNG format, which makes it challenging for AIGI detectors that leverage format shortcut. Moreover the generated images are also from popular latent diffusions, including DALL-E, Stable Diffusions, Firefly\cite{AdobeFirefly} and Glide.

\vspace{2mm}
\noindent
\textbf{Chameleon} \cite{yan2024sanity} provides a sanity check for AI-generated image detection. They build a high quality dataset where generated images are source from internet and some unknown source. All images in this benchmark are said to be indistinguishable by human. Since all images are gather from the unknown source, there's only one subset in this benchmark.

\vspace{2mm}
\noindent
\textbf{Community Forensics Evaluation Set }\cite{park2025commfor} is build to evaluate the model's ability to generalize to unseen generators that trained in Community Forensics dataset. This evaluation dataset is also the most up-to-date dataset, containing generators like Deci Diffusion V2~\cite{DeciFoundationModels}, GALIP~\cite{tao2023galip}, KandinskyV2.2~\cite{kandinsky2023models}, Kvikontent~\cite{KvikontentMidjourney}, LCM-LoRA-SDv1.5, LCM-LoRA-SDXL, LCM-LoRA-SSD1B~\cite{luo2023lcmLora}, Stable Cascade~\cite{pernias2023stablecascade}, DF-GAN~\cite{tao2022df}, and HDiT~\cite{crowson2024scalable}.

\begin{figure*}[!t]
        \centering
        \includegraphics[width=\linewidth]{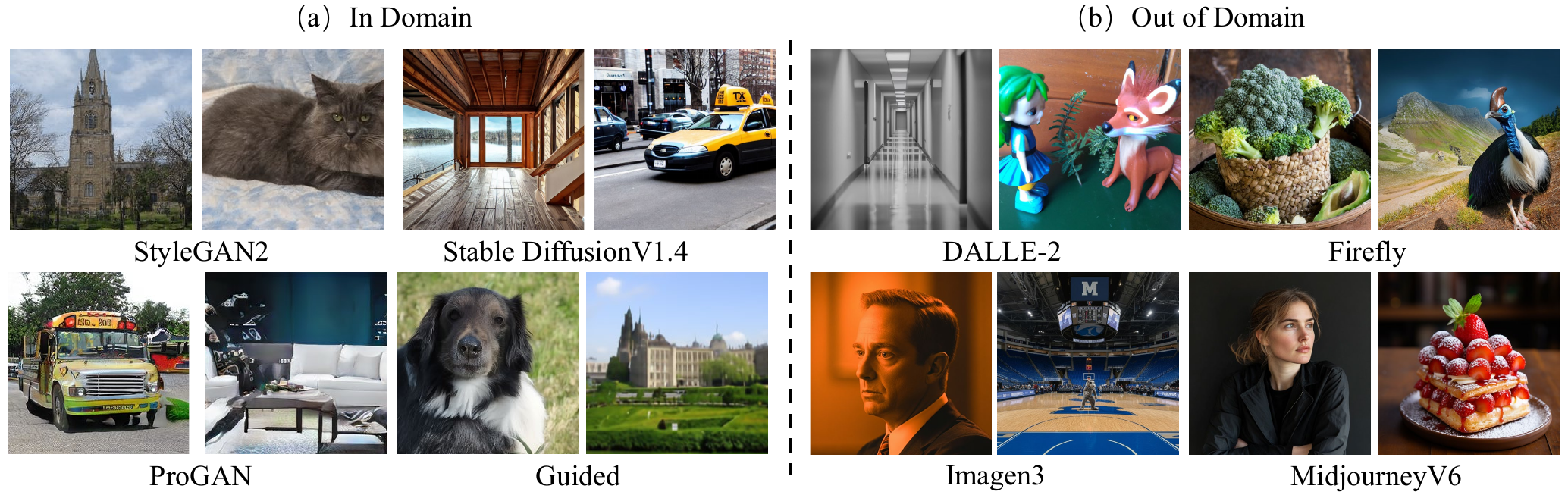} 
        \caption{Examples of test subsets, we visualize some in-domain datasets with our training set along with some out-of-distribution sets. }
        \label{fig:datasets} 
\end{figure*}

Above all, we have 55 subsets for testing. Given the large scale of our training data, the training domain overlaps with several previously constructed datasets. Consequently, our evaluation comprises $29$ completely unseen generator subsets, the rest, even though seen in training set, still have a different generated condition. Sample images from the test set are shown in Fig. ~\ref{fig:datasets}.

\textbf{Metrics.} Following prior works \cite{wang2020cnn, ojha2023towards, tan2024rethinking}, we compute a threshold-free metric, mean average precision (AP), and a threshold-based metric, binary accuracy (Acc). When computing accuracy, the threshold was set to 0.5.

\section{Baselines}
\label{sec:baseline}
    In this section, we will give a brief introduction to the baselines for comparison.

\subsection{GAN-Generalized}
\textbf{CNNDetection} \cite{wang2020cnn} uses a ResNet-50 as a classifier with data augmentation to detect CNN-generated images. \textbf{NPR} \cite{tan2024rethinking} rethink up-sampling operation in most generative architecture and detect them via a interpolation pattern. \textbf{UniFD} \cite{ojha2023towards} leverage the image encoder of CLIP for feature extraction, it takes image embeddings for classification with simple KNN or linear layer. \textbf{SAFE}\cite{li2025improving} extracts high frequency band as artifact with various data augmentation to build a CNN classifier. \textbf{AIDE} uses a hybrid feature of both CLIP image embedding , high and low frequency patches via DCT scoring, and concatenate them for final decision. These baselines are all trained on GAN generated images provided by the training set of CNNDetection \cite{wang2020cnn}. 

\subsection{Diffusion-Generalized}
In the diffusion era \cite{rombach2022high}, more method aim to detect generated images with a more realistic diffusion generated images. \textbf{DRCT} \cite{chen2024drct} construct a diffusion reconstruct dataset and used it to train a classifier with classification objective and contrastive learning objective. We use its Conv-B variant trained on SDv2.1. \textbf{Co-SPY} \cite{cheng2025co} also uses a hybrid feature for classification, but a combination of CLIP embedding and VAE-reconstruct residual. \textbf{B-Free} \cite{Guillaro2024biasfree} proposed a paradigm that generated images should come from inpainting models rather than unconditional generators or T2Is to prevent semantic bias. It trained an end-to-end classifier with proposed dataset. 

\subsection{Scaling-Ups}
Scaling up detectors use a training dataset from a more diverse source. To solve the sanity check for AIGI detection, AIDE \cite{yan2024sanity} provide another checkpoint that trained a classifier with all images in GenImage, including both GANs and Diffusions. D3 \cite{yang2025d3} proposed a dual feature extraction branch, a original image and a patch-shuffled image to learn comprehensive traces, it uses a training dataset consist of GenImage and CNNDetection. Community Forensics construct a large dataset with thousands of generators and use it to train classifier with a standard ViT.

\section{Detailed results of experiments}
The following tables show detailed results for the selected 6 benchmarks. All baselines are evaluated with their provided checkpoints. Tab.~\ref{tab:Forensynths} shows results of ForenSynths, Tab.~\ref{tab:UFD} shows results of UFD, Tab.~\ref{tab:GenImage} shows results of GenImage, Tab.~\ref{tab:Synthbuster} shows results of SynthBuster, Tab.~\ref{tab:commfor1}
and Tab.~\ref{tab:commfor2} shows results of Community Forensics evaluation set.

\label{sec:detailexp}
\section{Implementation Details}
\label{sec:impledet}
To validate the reproducibility, we present our implementation details in this section.

\textbf{Network Design} The image encoder we use is CLIP:ViT-L with a patch size of $14 \times 14$. The MLP is composed of two hidden layer with dimensions $1024 \xrightarrow{} 128 \xrightarrow{} 1$. In the second stage, we discard the second layer to make the MLP $1024 \xrightarrow{} 128$, then we apply LoRA \cite{hu2022lora} in \texttt{q\_proj, k\_proj, v\_proj} of image encoder with LoRA parameter $r=16, \alpha=32, \text{dropout}=0.1$.

\textbf{Training} For image in training set, we apply RandomCrop to $224 \times 224$, and to those image whose resolution is smaller than this, we apply 0-padding to make it enough to crop. We automatically separate 5\%  samples of dataset to form a validation set. We use AdamW \cite{loshchilov2017decoupled} optimizer with a learning rate of $10^{-4}$ and a weight decay of $0.01$ in the two stages. In the first stage, training last for 20 epochs. In the second stage. we train until the validation accuracy reaches 99.9\% or do not improve for 3 epochs to prevent performance degradation.

\textbf{Testing} For images in testing sets, we apply CenterCrop to $224 \times 224$, which is the same as most baselines, except for \textbf{B-Free}, who uses a resolution of $504 \times 504$ and \textbf{Co-SPY} uses a $384 \times 384$, to best excel their performance, we did not modify their model but directly report their performance on original resolution.

\section{Additional Derivation and Analysis}
\label{sec:additional_analysis}

In this section, we provide supplementary materials to substantiate the claims made in the main text. Specifically, we present a rigorous mathematical derivation modeling the data heterogeneity in scaling-up settings (Sec.~\ref{sec:theory_derivation}) and offer qualitative visualizations of the attention maps to demonstrate how our method alters the encoder's focus (Sec.~\ref{sec:vis_attention}).

\subsection{Detailed Derivation of Theory}
\label{sec:theory_derivation}

Recall that in Sec.~3 of the main paper, we identified the ``Benefit then Conflict'' dilemma, attributing it to the severe data-level heterogeneity that arises when aggregating multiple generators. To theoretically quantify this phenomenon, we model the distribution of real and generated images and analyze the behavior of feature variance.

\textbf{Why GMM? How to calculate variance in GMM?}

We adopt the Gaussian Mixture Model (GMM) as a proxy for the real-world data distribution. This choice is motivated by the structure of commonly used datasets like ImageNet, which consist of distinct categories. Specifically, we model the image features within each category as a multivariate Gaussian distribution. While this is a simplified assumption, it remains theoretically robust and yields conclusions consistent with more complex scenarios. The total variance of the generated distribution is derived as follows:
\begin{equation}
    \Sigma_{gen}  = \mathbb{E}_G \left[ \text{Var}(X |G) \right] + \text{Var} _G\left[ \mathbb{E}(X|G)\right],
\end{equation}
For the $i$-th generator, its variance is computed by:
\begin{equation}
\begin{split}
    & Var(X|G=i)=\Sigma_{gen}^i \\
    & = \sum_{j=1}^{M}\pi_{i,j}\Sigma_{i,j} + \sum_{j=1}^{M}\pi_{i,j}(\mu_{i,j}-\mu_i)(\mu_{i,j}-\mu_i)^T
\end{split}
\end{equation}
Thus, the total variance can be expressed as:
\begin{equation}
\begin{split}
    \Sigma_{gen}&=\sum_{i}^{G}w_i\sum_{j}^{M}\pi_{i,j}\Sigma_{i,j} \\
     & + \sum_{i}^{G}w_i\sum_{j}^{M}\pi_{i,j}(\mu_{i,j}-\bar{\mu})(\mu_{i,j}-\bar{\mu})^T \\
     & + \sum_{i}^{G}w_i(\mu_{i}-\bar{\mu})(\mu_{i}-\bar{\mu})^T \\
     & = \underbrace{\mathbb{E}_M \left[ \text{Var}(X|G_i,M)|G_i\right]+\text{Var}_M\left[\mathbb{E}(X|G_i,M)|G_i \right]}_{\text{generator fitting variance}} \\
        & + \underbrace{\text{Var}_G \left[ \mathbb{E}(X|G)\right]}_{\text{cross generator variance}},
\end{split}
\end{equation}
where $\mu_i=\sum_{j}^{M}\pi_{i,j}\mu_{i,j}$, and $\bar{\mu}=\sum_{i}^{G}\mu_i$ denotes the mean within a generator and the global mean across the entire dataset, respectively.

\textbf{How do prototypes limit variance growth?}

Having established that variance increases with generator diversity, we now show how our prototype mapping strategy mitigates this issue. In prototype mapping, the feature embedding is reorganized into a linear combination of prototypes; this operation effectively sets an upper bound on the embedding's variance.

We have $ \tilde{f}=\sum w_i v_i$ where $w_i$ is the attention score and $v_i$ is a prototype vector. Let $F$ be the random variable of this reorganized embedding. Its variance is given by:
\begin{equation}
    Var(F)=\mathbb{E} \left[ || F-\mathbb{E}(F)||^2\right],
\end{equation}
where $Var(\cdot)$ here denotes its trace in the covariance matrix, i.e., $tr(\Sigma)$. To proceed, we introduce an independent identically distributed random variable $F'$ to reformulate this term as:
\begin{equation}
    \begin{aligned}
        Var(F)& =\frac{1}{2}\mathbb{E} \left[ ||F - F'||^2 \right] \\
        & = \frac{1}{2} \sum w_i w_j|| v_i-v_j||^2,
    \end{aligned}
\end{equation}
Assume $ D= \max_{v \in P}||v_i-v_j||$. Since all the prototypes are determined and fixed, this maximum distance is a constant. Due to the fact that $w_i, w_j \in [0,1]$ and $\sum w_i = 1$, this variance term can be bounded by:
\begin{equation}
    Var(F) \le \frac{1}{4}D^2.
\end{equation}
This derivation confirms that regardless of the number of source generators, the variance of the features mapped to the prototype space remains bounded by the geometry of the prototype set.

\subsection{Visualization of Attention Maps}
\label{sec:vis_attention}

To provide interpretability for our model's performance, we move beyond theoretical bounds to empirical visualization. We compare the attention maps from both the original CLIP image encoder and the encoder fine-tuned by our proposed GAPL. 

\begin{figure*}[!t]
        \centering
        \includegraphics[width=0.85\linewidth]{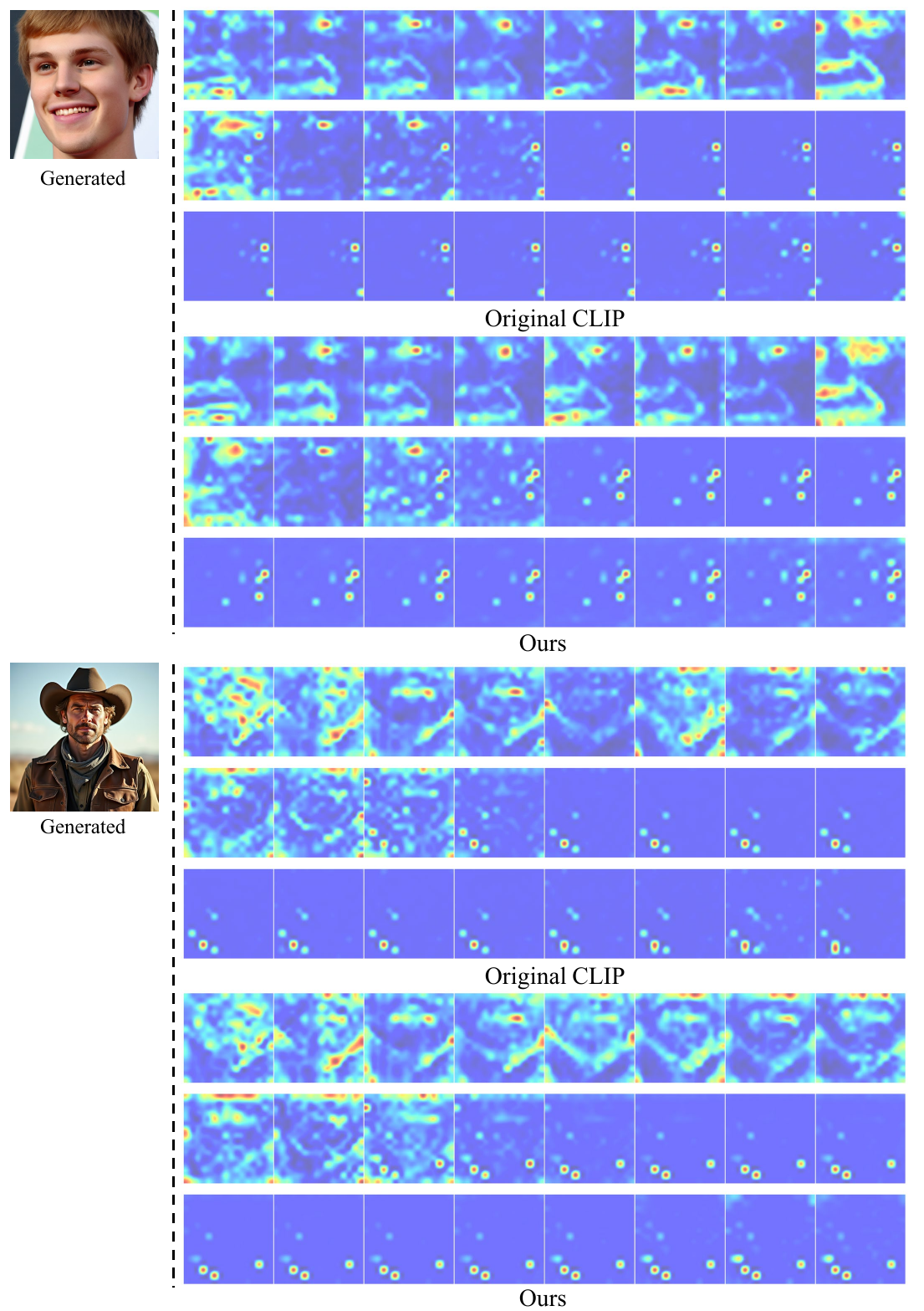} 
        \caption{\textbf{Self-Attention map} between original CLIP backbone and our finetuned backbone. There are 24 ViT blocks in the image encoder, we plot 8 blocks in each row, with indices increasing from left to right. For clarity of visualization, we use bicubic interpolation between image patches. In the shallow layers, we preserve most semantic features, in deep layers, our attention includes a wider range compared to original CLIP. }
        \label{fig:attention1} 
\end{figure*}

\begin{figure*}[!t]
        \centering
        \includegraphics[width=0.85\linewidth]{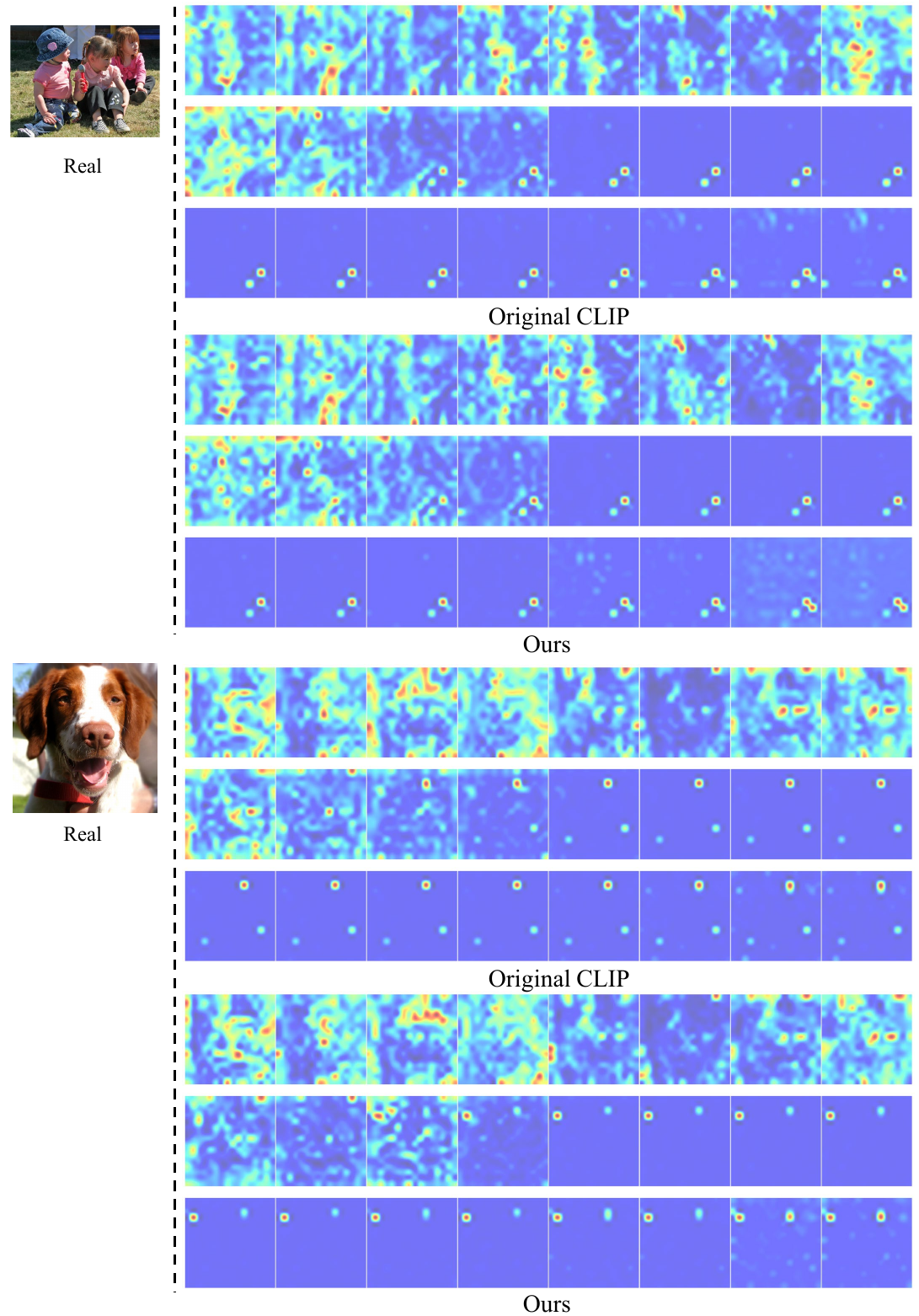} 
        \caption{\textbf{Self-Attention map} between original CLIP backbone and our finetuned backbone. There are 24 ViT blocks in the image encoder, we plot 8 blocks in each row, with indices increasing from left to right. For clarity of visualization, we use bicubic interpolation between image patches. In the shallow layers, we preserve most semantic features, in deep layers, our attention includes a wider range compared to original CLIP.}
        \label{fig:attention2} 
\end{figure*}

These attention maps represent the attention scores between the [CLS] token and all spatial patch tokens. The image encoder consists of 24 ViT blocks; in our visualization, we sample 8 blocks, with block indices ascending from left to right (shallow to deep). For visual clarity, bicubic interpolation is applied to the raw attention maps. As shown in Fig.~\ref{fig:attention1} and Fig.~\ref{fig:attention2}, the comparisons demonstrate two key findings: (1) Our encoder preserves rich semantic information in the shallow layers, maintaining the generalization capability of the pre-trained model; (2) In deeper layers, compared to the original CLIP, our model effectively leverages more patches to form a more comprehensive artifact pattern. Visually, this manifests as more pronounced and focused bright spots in the deeper layers, indicating that GAPL has successfully learned to attend to generator-specific traces.

\section{Future Perspectives}
\label{sec:future}

While our proposed GAPL framework achieves state-of-the-art performance by managing data heterogeneity through generator-aware prototypes, we recognize that the arms race between generation and detection is evolving. As generative models (e.g., Flux~\cite{blackforestlabs2024flux}, Midjourney v6~\cite{midjourneyv6}, Nano Banana Pro~\cite{banana}) increasingly mitigate low-level statistical artifacts, relying solely on texture or frequency cues may face diminishing returns. We envision that the next generation of AIGI detectors must integrate higher-level cognitive capabilities. Specifically, we identify three promising avenues where our scaling-up principles can intersect with broader AI domains:

\paragraph{From Artifact Detection to Visual Reasoning.}
Current forensic methods predominantly focus on signal-level anomalies. However, highly realistic AI-generated images often retain subtle \textit{semantic} or \textit{logical} inconsistencies (e.g., impossible shadows, mismatched reflections, or counting errors) that are invisible to standard classifiers. Future work should explore integrating \textbf{Visual Reasoning} frameworks~\cite{tanreason, ji2025mathsticks, ji2025visualtrans,song2025maniplvm} into the detection pipeline. By leveraging neuro-symbolic approaches or chain-of-thought reasoning, a detector could move beyond binary classification to provide interpretable evidence based on compositional logic, effectively identifying ``why'' an image violates reality even when pixel statistics appear perfect.

\paragraph{Incorporating Spatial Intelligence and Physical Constraints.}
A persistent weakness in current generative models is their frequent violation of 3D geometry and physical laws. While GAPL effectively learns 2D prototypes, it does not explicitly model the underlying 3D structure of the scene. Incorporating {spatial intelligence}~\cite{ji2025robobrain,team2025robobrain,wang2025crosspoint} could serve as a powerful orthogonal check. Models equipped with 3D-aware representations or spatial reasoning capabilities can detect geometric implausibilities—such as ``Escher-like'' structures or inconsistent depth cues—that purely 2D generative models fail to resolve. This aligns with our observation of ``Benefit then Conflict,'' suggesting that physical laws could serve as the ultimate invariant feature across diverse generators.

\paragraph{Trustworthy Perception in Embodied AI.}
Finally, the utility of AIGI detection extends beyond digital media forensics into the physical world. As we deploy autonomous agents, ensuring {trustworthy perception} is critical. Embodied agents operating in the real world must distinguish between authentic sensory inputs and potentially manipulated streams (e.g., adversarial projections or deepfakes in video feeds)~\cite{ji2025enhancing,bai2025alleviating}. Furthermore, robust detection models like GAPL can act as quality filters for {Embodied AI}~\cite{bai2025towards,lyu2025egoprompt,ji2025visualtrans,tan2025roboos}, particularly in Sim-to-Real pipelines where agents are trained on synthetic data. By rigorously curating high-fidelity synthetic data that adheres to physical realism, we can ensure that embodied agents develop robust representations that generalize better to physical reality.

\begin{table*}[t]
\centering
\renewcommand{\arraystretch}{1.1}
\resizebox{\textwidth}{!}{
\begin{tabular}{l cc cc cc cc cc cc cc cc cc}
\toprule
\multirow{2}{*}{\textbf{Method}} & \multicolumn{2}{c}{ProGAN} & \multicolumn{2}{c}{StyleGAN} & \multicolumn{2}{c}{StyleGAN2} & \multicolumn{2}{c}{StarGAN} & \multicolumn{2}{c}{GauGAN} & \multicolumn{2}{c}{CycleGAN} & \multicolumn{2}{c}{BigGAN} & \multicolumn{2}{c}{Deepfake} & \multicolumn{2}{c}{Mean} \\
\cmidrule(lr){2-3} \cmidrule(lr){4-5} \cmidrule(lr){6-7} \cmidrule(lr){8-9} \cmidrule(lr){10-11} \cmidrule(lr){12-13} \cmidrule(lr){14-15} \cmidrule(lr){16-17} \cmidrule(lr){18-19}
& Acc & AP &  Acc & AP & Acc & AP& Acc & AP&  Acc & AP & Acc & AP& Acc & AP & Acc & AP & Acc & AP \\
\midrule
CNNDet \cite{wang2020cnn} & 100.0 & 100.0 & 74.3 & 98.4 & 76.3 & 92.1 & 81.1 & 95.3 & 80.1 & 98.0 & 81.1 & 96.4 & 59.5 & 88.2 & 51.0 & 66.1 & 75.3 $\pm$ 13.9 & 91.8 $\pm$ 10.3 \\ 
NPR \cite{tan2024rethinking}  & 99.8 & 99.9 & 97.3 & 99.9 & 99.5 & 100 & 99.7 & 100 & 79.1 & 80.2 & 94.5 & 97.6 & 83.6 & 84.7 & 73.7 & 72.5 & 90.1 $\pm$ 9.9 & 91.8 $\pm$ 10.3 \\
UniFD \cite{ojha2023towards} &  99.8 & 100.0 & 85.4 & 97.5 & 74.5 & 97.6 & 96.0 & 99.5 & 99.4 & 100.0 & 98.3 & 99.8 & 94.7 & 99.2 & 67.5 & 80.3 & 89.4 $\pm$  11.6 & 96.7 $\pm$  6.3 \\
SAFE \cite{li2025improving} & 99.8 & 100.0 & 97.6 & 99.8 & 98.7 & 100.0 & 99.8 & 100.0 & 92.2 & 96.9 & 99.1 & 99.8 & 89.5 & 95.1 & 93.2 & 97.2 & 96.2  $\pm$ 3.7 & 98.6 $\pm$ 1.8 \\
AIDE \cite{yan2024sanity} & 96.3 & 99.8 & 97.2 & 99.7 & 98.1 & 99.8 & 98.4 & 99.9 & 76.9 & 96.0 & 95.6 & 99.4 & 78.3 & 96.7 & 54.2 & 68.7 & 86.8 $\pm$  14.9 & 95.0 $\pm$ 10.0 \\
DRCT \cite{chen2024drct} &  50.2 & 50.2 & 49.1 & 49.1 & 49.3 & 45.3 & 38.3 & 38.7 & 49.9 & 43.2 & 49.3 & 44.6 & 49.7 & 47.8 & 64.7 & 78.4 & 50.1 $\pm$ 6.7 & 49.6 $\pm$ 11.4 \\
Co-SPY \cite{cheng2025co} & 74.7 & 78.1 & 63.7 & 69.8 & 59.7 & 62.9 & 62.1 & 94.3 & 69.6 & 83.4 & 58.5 & 55.8 & 71.6 & 83.9 & 64.9 & 78.7 & 65.6 $\pm$ 5.4 & 75.9 $\pm$ 11.6 \\
B-Free \cite{Guillaro2024biasfree} & 95.6 & 99.3 & 74.8 & 93.6 & 71.1 & 89.4 & 81.1 & 93.5 & 96.0 & 99.8 & 65.4 & 90.5 & 91.5 & 98.9 & 73.7 & 89.2 & 81.2 $\pm$  11.1 & 94.3 $\pm$ 4.2  \\ 
AIDE$\dagger$ \cite{yan2024sanity} & 92.7 & 98.8 & 88.6 & 95.1 & 93.6 & 98.6 & 88.0 & 99.0 & 87.7 & 98.9 & 92.0 & 98.9 & 84.9 & 97.7 & 54.8 & 63.2 & 85.3 $\pm$  11.8 & 93.8 $\pm$ 11.6 \\
D3 \cite{yang2025d3} & 99.8 & 100.0 & 93.5 & 99.1 & 95.8 & 99.5 & 93.5 & 98.7 & 98.7 & 100.0 & 95.9 & 99.5 & 99.6 & 100.0 & 67.5 & 87.0 & 93.0 $\pm$ 9.9 & 98.0 $\pm$ 4.2 \\
CommForen \cite{park2025commfor} & 92.8 & 99.8 & 93.1 & 99.3 & 92.7 & 99.5 &  98.8 & 99.9 & 98.8 & 99.9 & 95.5 & 99.8 & 99.6 & 100.0 & 66.8 & 88.9 & 92.2 $\pm$ 10.0 & 98.2 $\pm$ 3.6 \\
\rowcolor{cvprblue!20} 
GAPL(Ours) & 99.9 & 100.0 & 98.1 & 100.0 & 99.5 & 100.0 & 97.0 & 99.9 & 99.6 & 100.0 & 98.2 & 99.9 & 98.6 & 100.0 & 88.2 & 96.8 & 97.2 $\pm$ 3.6 & 99.5 $\pm$ 1.1 \\
\bottomrule
\end{tabular}
}
\caption{Detaild results on the benchmark ForenSynth \cite{wang2020cnn}. we only select its GAN variant. $\text{AIDE}^{\dagger}$ denotes its scaling up checkpoint trained on 8 generators.}
\label{tab:Forensynths}
\end{table*}

\begin{table*}[!t]
\centering
\renewcommand{\arraystretch}{1.1}
\resizebox{\textwidth}{!}{
\begin{tabular}{l cc cc cc cc cc cc cc cc cc}
\toprule
\multirow{2}{*}{\textbf{Method}} & \multicolumn{2}{c}{DALL-E} & \multicolumn{2}{c}{Glide-50-27} & \multicolumn{2}{c}{Glide-100-10} & \multicolumn{2}{c}{Glide-100-27} & \multicolumn{2}{c}{Guided} & \multicolumn{2}{c}{LDM-100} & \multicolumn{2}{c}{LDM-200} & \multicolumn{2}{c}{LDM-200-cfg} & \multicolumn{2}{c}{Mean} \\
\cmidrule(lr){2-3} \cmidrule(lr){4-5} \cmidrule(lr){6-7} \cmidrule(lr){8-9} \cmidrule(lr){10-11} \cmidrule(lr){12-13} \cmidrule(lr){14-15} \cmidrule(lr){16-17} \cmidrule(lr){18-19}
& Acc & AP &  Acc & AP & Acc & AP& Acc & AP&  Acc & AP & Acc & AP& Acc & AP & Acc & AP & Acc & AP \\
\midrule
CNNDet \cite{wang2020cnn} & 52.9 & 68.4 & 55.7 & 78.0 & 54.3 & 74.3 & 53.3 & 73.7 & 52.8 & 68.1 & 52.0 & 68.7 & 51.5 & 68.2 & 52.2 & 69.6 & 53.1 $\pm$ 1.3 & 71.1 $\pm$3.5  \\ 
NPR \cite{tan2024rethinking}  & 90.3 & 97.4 & 97.2 & 99.2 & 97.6 & 99.2 & 96.9 & 99.1 & 87.1 & 92.6 & 97.4 & 99.1 & 97.6 & 99.2 & 97.4 & 99.0 & 95.2 $\pm$ 3.8 & 98.1 $\pm$ 2.2 \\
UniFD \cite{ojha2023towards} & 87.5 & 97.7 & 79.2 & 96.0 & 78.0 & 95.5 & 78.7 & 95.8 & 70.0 & 88.3 & 95.2 & 99.3 & 94.5 & 99.4 & 74.2 & 93.2 & 82.2 $\pm$ 8.7 & 95.7$\pm$ 3.4 \\
SAFE \cite{li2025improving} & 97.5 & 99.7 & 96.6 & 99.2 & 97.3 & 99.4 & 95.8 & 98.9 & 82.4 & 95.8 & 98.8 & 100.0 & 98.8 & 100.0 & 98.7 & 99.9 & 95.7 $\pm$5.1 & 99.1 $\pm$ 1.3 \\
AIDE \cite{yan2024sanity} & 89.7 & 99.6 & 89.9 & 99.8 & 89.8 & 99.6 & 89.9 & 99.8 & 94.2 & 98.9 & 90.1 & 99.9 & 90.1 & 99.9 & 90.1 & 99.9 & 90.5 $\pm$1.4 & 99.7 $\pm$ 0.3 \\
DRCT \cite{chen2024drct} & 55.6 & 57.8 & 56.2 & 62.6 & 61.0 & 70.4 & 56.2 & 62.7 & 62.6 & 89.3 & 88.8 & 96.6 & 88.9 & 96.7 & 90.3 & 97.5 & 70.0 $\pm$ 15.2 & 79.2 $\pm$16.3  \\
Co-SPY \cite{cheng2025co} & 81.8 & 87.2 & 69.0 & 74.6 & 76.7 & 81.7 & 73.5 & 78.2 & 62.7 & 87.4 & 82.7 & 86.9 & 83.1 & 87.5 & 85.2 & 91.0 & 76.8 $\pm$7.4 & 84.3 $\pm$5.2 \\
B-Free \cite{Guillaro2024biasfree} & 93.2 & 97.9 & 78.6 & 90.2 & 81.8 & 91.9 & 77.9 & 89.4 & 74.6 & 93.6 & 97.2 & 99.9 & 97.1 & 99.8 & 96.9 & 99.8 & 87.1 $\pm$9.2 & 95.3 $\pm$ 4.2 \\ 
AIDE$^\dagger$ \cite{yan2024sanity} & 98.7 & 99.9 & 99.0 & 100.0 & 99.1 & 100.0 & 99.0 & 100.0 & 95.5 & 99.9 & 99.1 & 100.0 & 99.1 & 100.0 & 99.1 & 100.0 & 98.6 $\pm$ 1.1 & 100.0 $\pm$ 0 \\
D3 \cite{yang2025d3} & 94.0 & 98.7 & 95.8 & 99.3 & 95.8 & 99.4 & 95.7 & 99.5 & 95.9 & 99.6 & 96.1 & 99.7 & 95.8 & 99.7 & 89.3 & 96.5 & 94.8 $\pm$ 2.2& 99.0 $\pm$ 1.0 \\
CommForen \cite{park2025commfor} & 98.4 & 99.9 & 97.1 & 99.6 & 98.2 & 99.8 & 96.8 & 99.6 & 66.1 & 76.6 & 98.5 & 99.9 & 98.7 & 99.9 & 98.5 & 99.9 & 94.0 $\pm$ 10.5 & 96.9 7.6 \\
\rowcolor{cvprblue!20} 
GAPL(Ours) & 97.8 & 100.0 & 97.7 & 99.9 & 97.8 & 100 & 97.7 & 100.0 &  93.4 & 98.6 &  97.8 & 100.0 & 97.9 & 100.0 & 97.9 & 100 & 97.2 $\pm$ 1.5  & 99.8 $\pm$ 0.5 \\
\bottomrule
\end{tabular}
}
\caption{Detaild results on the benchmark UFD. We refer this set to the diffusion part that \cite{ojha2023towards} added. }
\label{tab:UFD}
\end{table*}

\begin{table*}[!t]
\centering
\renewcommand{\arraystretch}{1.1}
\resizebox{\textwidth}{!}{
\begin{tabular}{l cc cc cc cc cc cc cc cc cc}
\toprule
\multirow{2}{*}{\textbf{Method}} & \multicolumn{2}{c}{VQDM} & \multicolumn{2}{c}{SDv1.4} & \multicolumn{2}{c}{BigGAN} & \multicolumn{2}{c}{Wukong} & \multicolumn{2}{c}{SDv1.5} & \multicolumn{2}{c}{Glide} & \multicolumn{2}{c}{Midjourney} & \multicolumn{2}{c}{ADM} & \multicolumn{2}{c}{Mean} \\
\cmidrule(lr){2-3} \cmidrule(lr){4-5} \cmidrule(lr){6-7} \cmidrule(lr){8-9} \cmidrule(lr){10-11} \cmidrule(lr){12-13} \cmidrule(lr){14-15} \cmidrule(lr){16-17} \cmidrule(lr){18-19}
& Acc & AP &  Acc & AP & Acc & AP& Acc & AP&  Acc & AP & Acc & AP& Acc & AP & Acc & AP & Acc & AP \\
\midrule
CNNDet \cite{wang2020cnn} & 52.1 & 68.8 & 50.5 & 59.8 & 51.2 & 78.6 & 50.8 & 59.3 & 50.7 & 60.6 & 52.3 & 67.6 & 51.3 & 64.7 & 51.5 & 65.6 & 51.3 $\pm$  0.6 & 65.6 $\pm$ 5.9  \\ 
NPR \cite{tan2024rethinking}  & 89.6 & 94.1 & 92.1 & 95.0 & 70.1 & 80.3 & 88.0 & 93.0 & 91.7 & 95.4 & 93.2 & 96.5 & 79.7 & 86.5 & 85.7 & 91.7 & 86.2 $\pm$ 7.3 & 91.6 $\pm$ 
 5.1 \\
UniFD \cite{ojha2023towards} & 84.0 & 96.3 & 64.1 & 87.4 & 89.8 & 98.5 & 71.8 & 91.2 & 64.4 & 86.0 & 61.6 & 84.6 & 57.2 & 75.5 & 67.1 & 86.8 & 70.0 $\pm$ 10.6 & 88.3 $\pm$  6.7 \\
SAFE \cite{li2025improving} & 96.1 & 99.6 & 99.4 & 100.0 & 97.6 & 99.8 & 98.1 & 99.8 & 98.8 & 99.9 & 97.2 & 99.6 & 95.5 & 99.5 & 81.5 & 96.4 & 95.5 $\pm$ 5.4 & 99.3 $\pm$ 1.1\\
AIDE \cite{yan2024sanity} & 98.4 & 99.9 & 98.3 & 99.9 & 98.9 & 99.9 & 97.3 & 99.8 & 98.1 & 99.8 & 98.6 & 99.9 & 86.4 & 96.6 & 96.8 & 99.5 & 96.6 $\pm$ 3.9 & 99.4 $\pm$ 1.1  \\
DRCT \cite{chen2024drct} & 61.3 & 85.4 & 97.9 & 99.8 & 52.3 & 86.3 & 95.9 & 99.6 & 97.9 & 99.8 & 60.1 & 91.2 & 98.2 & 99.9 & 60.0 & 90.2 & 78.0 $\pm$ 20.0 & 94.0 $\pm$  6.0 \\
Co-SPY \cite{cheng2025co} & 72.2 & 93.9 & 93.1 & 99.2 & 64.7 & 90.8 & 91.2 & 98.8 & 93.0 & 99.0 & 81.4 & 96.5 & 70.0 & 91.0 & 58.8 & 84.3 & 78.0 $\pm$  12.7 & 94.2 $\pm$  4.9 \\
B-Free \cite{Guillaro2024biasfree} & 88.2 & 98.3 & 99.4 & 100.0 & 76.5 & 96.8 & 99.4 & 100.0 & 99.3 & 100.0 & 69.7 & 93.8 & 94.1 & 99.3 & 72.5 & 93.7 & 87.4 $\pm$ 11.9 & 97.7 $\pm$  2.5 \\ 
AIDE$^\dagger$ \cite{yan2024sanity} & 99.9 & 100.0 & 99.8 & 100.0 & 99.9 & 100.0 & 99.6 & 100.0 & 99.8 & 100.0 & 99.8 & 100.0 & 99.0 & 100.0 & 99.5 & 100.0 & 99.7 $\pm$ 
 0.2 & 100.0$\pm$  0\\
D3 \cite{yang2025d3} & 98.3 & 99.9 & 98.1 & 99.8 & 97.1 & 99.7 & 97.7 & 99.8 & 97.7 & 99.8 & 98.3 & 99.8 & 79.7 & 95.8 & 96.8 & 99.6 & 95.5 $\pm$  6.0  & 99.3 $\pm$ 1.3  \\
CommForen \cite{park2025commfor} & 90.9 & 99.7 & 90.6 & 99.2 & 71.5 & 81.6 & 91.0 & 99.1 & 90.5 & 99.2 & 89.4 & 98.4 & 77.5 & 86.8 & 72.9 & 83.5 & 84.3 $\pm$ 8.1 & 93.4 $\pm$ 7.5\\
\rowcolor{cvprblue!20} 
GAPL(Ours) & 98.2 & 100 & 98.2 & 100 & 97.9 & 99.8 &  98.1 & 99.9 & 98.0 & 99.9 & 98.1 & 99.9 & 90.3 & 97.6 & 95.0 & 99.2 & 96.7$\pm$ 2.6 & 99.6 $\pm$ 0.7 \\
\bottomrule
\end{tabular}
}
\caption{Detaild results on the benchmark GenImage.}
\label{tab:GenImage}
\end{table*}

\clearpage

\begin{table*}[!t]
\centering
\renewcommand{\arraystretch}{1.1}
\resizebox{\textwidth}{!}{
\begin{tabular}{l c| cc cc cc cc cc cc cc cc cc cc}
\toprule
\multirow{2}{*}{\textbf{Method}} & \multicolumn{1}{c|}{Real} & \multicolumn{2}{c}{DALLE-2} & \multicolumn{2}{c}{Firefly} & \multicolumn{2}{c}{SDv1.4} & \multicolumn{2}{c}{SDXL} & \multicolumn{2}{c}{DALLE-3} & \multicolumn{2}{c}{Glide} & \multicolumn{2}{c}{MJ-v5} &\multicolumn{2}{c}{SDv1.3} & \multicolumn{2}{c}{SDv2} & \multicolumn{2}{c}{Mean} \\

\cmidrule(lr){2-2} \cmidrule(lr){3-4} \cmidrule(lr){5-6} \cmidrule(lr){7-8} \cmidrule(lr){9-10} \cmidrule(lr){11-12} \cmidrule(lr){13-14} \cmidrule(lr){15-16} \cmidrule(lr){17-18} \cmidrule(lr){19-20} \cmidrule(lr){21-22}
& Acc & Acc & AP & Acc & AP & Acc & AP& Acc & AP& Acc & AP & Acc & AP& Acc & AP & Acc & AP & Acc & AP & Acc & AP \\
\midrule
CNNDet \cite{wang2020cnn} & 96.5 & 7.0 & 64.8 & 14.4 & 74.9 & 1.5 & 49.9 & 5.2 & 59.0 & 0.1 & 36.2 & 1.8 & 46.9 & 1.2 & 44.8 & 1.3 & 49.3 & 4.8 & 96.5 & 50.3 $\pm$ 2.1 & 54.0 $\pm$ 17.9\\ 
NPR \cite{tan2024rethinking}  & 5.6 & 98.4 & 49.6 & 5.2 & 33.3 & 94.3 & 50.6 & 100 & 54.7 & 15.5 & 31.0 & 91.6 & 50.8 & 80.5 & 46.1 & 94.3 & 50.0 & 70.4 & 42.7 & 38.9 $\pm$17.1  & 45.4$\pm$7.7 \\
UniFD \cite{ojha2023towards} & 93.5 & 77.0 & 95.2 & 86.0 & 97.4 & 45.6 & 85.3 & 41.4 & 83.5 & 0.7 & 41.2 & 11.5 & 63.5 & 10.4 & 61.0 &44.8 & 85.3 &58.6 & 89.9 & 67.6 $\pm$14.0 & 78.0 $\pm$ 17.6 \\
SAFE \cite{li2025improving} & 15.5 & 92.0 & 42.2 & 4.3 & 31.0 & 91.8 & 55.9 & 87.4 & 37.1 & 44.5 & 34.7 & 58.1 & 36.0 & 97.0 & 54.8 & 91.0 & 53.9 & 99.5 & 56.5 & 44.7$\pm$15.1 & 44.7 $\pm$ 9.8\\
AIDE \cite{yan2024sanity} & 66.7 & 45.7 & 57.0 & 0.0 & 30.8 & 95.1 & 94.8 & 98.2 & 94.3 & 2.8 & 35.9 & 96.7 & 95.7 & 75.3 & 80.3 & 95.8 & 95.3 & 95.6 & 93.3 & 67.0 $\pm$ 19.3 & 75.3 $\pm$25.3  \\
DRCT \cite{chen2024drct} & 96.1 & 4.1 & 53.6 & 11.4 & 60.7 & 88.2 & 97.9 & 89.6 & 98.2 & 35.6 & 80.8 &  14.1 & 72.9 & 99.4 & 99.9 & 89.6 & 98.2 & 99.9 & 100.0 & 77.6 $\pm$ 19.6 & 84.7$\pm$ 17.3 \\
Co-SPY \cite{cheng2025co} & 97.6 & 48.5 & 90.8 & 43.6 & 87.5 & 74.7 & 96.4 & 44.1 & 87.9 & 73.7 & 96.5 & 80.6 & 97.8 & 35.2 & 85.0 & 74.5 & 96.7 & 40.0 & 85.4 & 77.4 $\pm$ 8.6  & 91.6 $\pm$ 5.0 \\
B-Free \cite{Guillaro2024biasfree} & 98.5 & 88.5 & 98.9 & 99.2 & 99.9 & 99.8 & 99.9 & 100.0 & 100.0 & 93.1 & 99.4 & 42.7 & 91.8 & 98.9 & 99.9 & 100.0 & 100.0 & 99.5 & 99.9 & 94.9$\pm$ 8.8  & 98.9 $\pm$ 2.5 \\ 
AIDE$^\dagger$ \cite{yan2024sanity} & 32.8 & 27.3 & 35.8 & 0.3 & 31.5 & 99.0 & 64.4 & 98.6 & 64.8 & 34.8 & 43.4 & 97.1 & 66.8 & 98.6 & 75.0 & 99.7 & 65.8 & 96.7 & 57.0 & 52.7 $\pm$ 18.8& 56.0 $\pm$ 14.5 \\
D3 \cite{yang2025d3} & 82.0 & 87.7 & 92.6 & 92.0 & 95.3 & 96.5 & 97.9 & 89.4 & 93.6 & 28.6 & 61.3 & 90.6 & 95.3 & 62.6 &81.2 & 96.4 & 97.7 & 82.6 & 90.8 & 81.3 $\pm$ 10.4 & 89.5  $\pm$ 11.0 \\
CommForen \cite{park2025commfor} & 84.6 & 83.9 & 91.6 & 93.1 & 96.6 & 94.7 & 97.4 & 98.5 & 98.3 & 76.1 & 88.4 & 90.0 & 96.0 & 80.6 & 91.4 & 95.7 & 97.7 & 92.9 & 95.3 & 87.1 $\pm$3.6 & 94.8 $\pm$ 3.3 \\
\rowcolor{cvprblue!20} 
GAPL(Ours) & 90.0 & 94.0 & 97.5 & 94.3 & 98.2 & 98.1 & 99.3 & 99.8 & 99.8 & 60.1 & 85.3 & 97.6 & 99.2 & 92.0 & 97.4 & 98.2 & 99.3 & 96.4 & 98.7 & 91.1 $\pm$ 5.8 & 97.2 $\pm$4.3 \\
\bottomrule
\end{tabular}
}
\caption{Detailed results on the benchmark SynthBuster \cite{synthbuster}. In this benchmark, all generated images are pair with exactly the same real images. Thus we report the real images accuracy and each generator's fake accuracy independently. We pair the metrics of real and each generators to get the final mean metrics .}
\label{tab:Synthbuster}
\end{table*}

\begin{table*}[!t]
\centering
\renewcommand{\arraystretch}{1.1}
\resizebox{\textwidth}{!}{
\begin{tabular}{l cc cc cc cc cc cc cc cc cc cc cc}
\toprule
\multirow{2}{*}{\textbf{Method}} & \multicolumn{2}{c}{DFGAN} & \multicolumn{2}{c}{MJv6} & \multicolumn{2}{c}{Kandinsky} & \multicolumn{2}{c}{SDcas.} & \multicolumn{2}{c}{MJv5} & \multicolumn{2}{c}{Firefly2} & \multicolumn{2}{c}{Firefly3} & \multicolumn{2}{c}{GALIP} & \multicolumn{2}{c}{LCM-lora-sdxl} & \multicolumn{2}{c}{Hourglass} & \multicolumn{2}{c}{Kvikontent} \\
\cmidrule(lr){2-3} \cmidrule(lr){4-5} \cmidrule(lr){6-7} \cmidrule(lr){8-9} \cmidrule(lr){10-11} \cmidrule(lr){12-13} \cmidrule(lr){14-15} \cmidrule(lr){16-17} \cmidrule(lr){18-19} \cmidrule(lr){20-21} \cmidrule(lr){22-23} 
& Acc & AP &  Acc & AP & Acc & AP& Acc & AP&  Acc & AP & Acc & AP& Acc & AP & Acc & AP & Acc & AP  & Acc & AP & Acc & AP\\
\midrule
NPR \cite{tan2024rethinking}  & 96.6 & 100.0 & 98.6 & 99.0 & 57.7 & 56.1 & 57.4 & 56.4 & 96.7 & 98.3 & 49.2 & 45.0 & 49.6 & 52.4 & 51.7 & 48.0 & 39.9 & 41.6 & 88.6 & 91.6 & 58.6 & 56.3 \\
SAFE \cite{li2025improving} & 50.3 & 44.8 & 49.9 & 45.0 & 49.6 & 53.2 & 49.9 & 51.0 & 49.9 & 45.1 & 50.0 & 51.3 & 50.1 & 51.2 & 49.5 & 53.9 & 49.5 & 52.3 & 50.1 & 45.2 & 50.2 & 51.5 \\
AIDE \cite{yan2024sanity} & 49.9 & 43.9 & 49.8 & 44.7 & 49.9 & 50.1 & 50.2 & 52.6 & 49.9 & 44.4 & 50.2 & 46.2 & 50.2 & 46.8 & 49.7 & 53.0 & 49.7 & 46.1 & 49.8 & 46.1 & 50.1 & 54.1 \\
DRCT \cite{chen2024drct} & 49.9 & 45.9 & 49.6 & 47.0 & 49.6 & 53.1 & 49.4 & 52.7 & 49.7 & 50.5 & 49.6 & 49.1 & 49.9 & 47.0 & 50.2 & 56.6 & 50.0 & 49.0 & 50.0 & 46.9 & 49.6 & 52.2 \\
Co-SPY \cite{cheng2025co} & 50.0 & 84.5 & 70.1 & 86.5 & 68.4 & 74.8 & 71.0 & 76.7 & 63.0 & 80.8 & 68.6 & 86.4 & 86.7 & 95.8 & 38.4 & 34.1 & 43.1 & 38.8 & 52.8 & 59.1 & 74.4 & 84.8  \\
B-Free \cite{Guillaro2024biasfree} & 92.2 & 96.5 & 76.7 & 84.9 & 85.7 & 99.4 & 86.1 & 99.3 & 86.2 & 97.2 & 82.9 & 92.7 & 81.8 & 91.7 & 81.8 & 91.9 & 85.0 & 97.6 & 57.6 & 65.6 & 86.6 & 99.5 \\ 
AIDE$^\dagger$ \cite{yan2024sanity} & 49.9 & 56.4 & 49.7 & 46.9 & 50.4 & 53.3 & 50.8 & 53.8 & 50.2 & 47.5 & 49.7 & 50.7 & 50.0 & 47.8 & 50.4 & 55.9 & 50.4 & 51.8 & 50.0 & 44.7 & 50.3 & 48.1 \\
D3 \cite{yang2025d3} & 98.4 & 100.0 & 69.5 & 81.0 & 67.9 & 84.5 & 71.5 & 90.0 & 76.7 & 87.0 & 80.0 & 90.8 & 78.6 & 88.7 & 70.8 & 90.8 & 63.0 & 77.7 & 68.3 & 83.1 & 74.9 & 95.7 \\
\rowcolor{cvprblue!20} 
GAPL(Ours) & 99.0 & 100 & 87.1 & 97.3 & 93.0 & 99.1 & 93.3 & 99.6 & 88.6 & 99.3 & 87.3 & 96.2 & 85.1 & 93.7 & 90.6 & 98.1 & 88.2 & 95.8 & 78.5 & 89.4 & 94.5 & 99.8  \\
\bottomrule 
\end{tabular}
}
\caption{Detailed results on the benchmark Community forensic evaluation. This is the first part. Note that the results of CNNDet, UniFD and CommForen are directly cited from the original paper \cite{park2025commfor} and dataset repository, whose detailed results are not available.}
\label{tab:commfor1}
\end{table*}

\begin{table*}[!t]
\centering
\renewcommand{\arraystretch}{1.1}
\resizebox{\textwidth}{!}{
\begin{tabular}{l cc cc cc cc cc cc cc cc cc cc cc}
\toprule
\multirow{2}{*}{\textbf{Method}} & \multicolumn{2}{c}{DALL-E 2} & \multicolumn{2}{c}{DALL-E 3} & \multicolumn{2}{c}{LCM-lora-sdv1.5} & \multicolumn{2}{c}{DeciDiff.} & \multicolumn{2}{c}{FLUX-dev} & \multicolumn{2}{c}{FLUX-schnell} & \multicolumn{2}{c}{IdeogramV2} & \multicolumn{2}{c}{IDeogramV1} & \multicolumn{2}{c}{Imagen3} & \multicolumn{2}{c}{LCM-lora-ssd1b} & \multicolumn{2}{c}{Mean} \\
\cmidrule(lr){2-3} \cmidrule(lr){4-5} \cmidrule(lr){6-7} \cmidrule(lr){8-9} \cmidrule(lr){10-11} \cmidrule(lr){12-13} \cmidrule(lr){14-15} \cmidrule(lr){16-17} \cmidrule(lr){18-19} \cmidrule(lr){20-21} \cmidrule(lr){22-23} 
& Acc & AP &  Acc & AP & Acc & AP& Acc & AP&  Acc & AP & Acc & AP& Acc & AP & Acc & AP & Acc & AP  & Acc & AP & Acc & AP\\
\midrule
NPR \cite{tan2024rethinking}  & 91.7 & 97.6 & 96.5 & 99.2 & 53.6 & 50.1 & 43.9 & 44.1 & 98.1 & 98.9 & 97.8 & 99.0 & 96.4 & 98.8 & 96.5 & 98.5 & 98.6 & 98.7 & 32.3 & 38.4 & 73.8 $\pm$ 24.0 & 74.7 $\pm$  25.0 \\
SAFE \cite{li2025improving} & 50.7 & 44.8 & 50.1 & 45.1 & 49.7 & 50.8 & 49.7 & 50.5 & 49.9 & 45.1 & 50.1 & 45.0 & 50.1 & 45.0 & 49.9 & 45.1 & 50.1 & 44.9 & 50.0 & 53.2 & 50.0 $\pm$ 0.3 & 48.3  $\pm$ 3.5 \\
AIDE \cite{yan2024sanity} & 49.6 & 44.2 & 50.0 & 45.2 & 50.1 & 52.4 & 50.0 & 48.5 & 49.9 & 44.9 & 49.8 & 44.5 & 50.0 & 44.4 & 49.9 & 44.7 & 49.8 & 45.0 & 50.0 & 47.7 & 49.9 $\pm$ 0.2 & 47.1 $\pm$ 3.2 \\
DRCT \cite{chen2024drct} & 49.3 & 46.8 & 49.5 & 49.0 & 49.8 & 51.5 & 49.6 & 51.8 & 48.4 & 46.8 & 48.6 & 47.0 & 50.0 & 49.1 & 49.6 & 49.4 & 49.0 & 49.0 & 49.2 & 46.9 & 49.5 $\pm$ 0.4 & 49.4 $\pm$ 0.7 \\
Co-SPY \cite{cheng2025co} & 77.4 & 91.3 & 85.2 & 96.0 & 67.2 & 74.1 & 71.7 & 79.8 & 80.1 & 93.5 & 71.8 & 87.6 & 71.2 & 88.3 & 75.6 & 90.8 & 77.0 & 91.8 & 61.0 & 68.3 & 67.9 $\pm$ 12.4 & 79.2 $\pm$ 16.6 \\
B-Free \cite{Guillaro2024biasfree} & 74.3 & 81.3 & 86.5 & 98.9 & 85.5 & 99.0 & 86.0 & 99.3 & 73.8 & 82.7 & 79.0 & 88.2 & 81.2 & 89.2 & 81.8 & 91.4 & 74.6 & 83.0 & 87.0 & 98.2 & 81.5 $\pm$ 7.1 & 91.8 $\pm$ 8.4  \\ 
AIDE$^\dagger$ \cite{yan2024sanity} & 49.3 & 47.4 & 49.9 & 45.1 & 50.6 & 48.2 & 50.4 & 50.7 & 49.9 & 40.9 & 50.1 & 41.7 & 50.0 & 42.9 & 50.0 & 43.7 & 50.1 & 44.4 & 50.0 & 51.8 & 50.1 $\pm$ 0.3 & 48.3 $\pm$ 4.4 \\
D3 \cite{yang2025d3} & 87.5 & 95.0 & 90.1 & 96.6 & 67.0 & 80.6 & 62.1 & 76.4 & 79.3 & 89.7 & 73.5 & 83.8 & 67.2 & 77.5 & 67.6 & 78.6 & 70.7 & 83.1 & 61.5 & 74.5 & 73.6 $\pm$ 9.3 & 86.0 7.1 \\
\rowcolor{cvprblue!20} 
GAPL(Ours) & 88.4 & 97.5 & 88.9 & 99.6 & 93.9 & 99.7 & 92.3 & 99.1 & 87.3 & 98.9 & 88.0 & 98.2 & 88.0 & 98.2 & 87.7 & 98.7 & 87.0 & 95.9 & 90.4 & 97.6 & 89.4$\pm$  4.0 & 97.8$\pm$ 2.4 \\
\bottomrule
\end{tabular}
}
\caption{Detailed results on the benchmark Community forensic evaluation. This is the second part.}
\label{tab:commfor2}
\end{table*}

\clearpage

\end{document}